\documentclass{article}

     \PassOptionsToPackage{numbers, compress}{natbib}
     \usepackage[final]{neurips_2019}

\usepackage[utf8]{inputenc} %
\usepackage[T1]{fontenc}    %
\usepackage{hyperref}       %
\usepackage{url}            %
\usepackage{booktabs}       %
\usepackage{amsfonts}       %
\usepackage{nicefrac}       %
\usepackage{microtype}      %

\usepackage{graphicx}
\usepackage{overpic}
\usepackage{amsmath}
\usepackage{amssymb}
\usepackage{mathtools}

\usepackage{wrapfig}
\usepackage{cleveref}
\usepackage{enumerate}
\usepackage{caption}

\usepackage{multirow}
\usepackage{xspace}

\DeclareMathOperator{\argmax}{argmax}
\DeclareMathOperator{\argmin}{argmin}
\DeclareMathOperator{\supp}{supp}
\DeclareMathOperator{\sgn}{sgn}

\DeclarePairedDelimiter\floor{\lfloor}{\rfloor}

\newcommand{\X}{\mathcal{X}}
\newcommand{\Y}{\mathcal{Y}}
\newcommand{\V}{\mathcal{V}}
\newcommand{\cP}{\mathcal{P}}
\newcommand{\dP}{\mathrm{d}\cP}
\newcommand{\E}{\mathbb{E}}
\newcommand{\Var}{\mathbb{V}}
\newcommand{\R}{\mathbb{R}}
\renewcommand{\P}{\mathbb{P}}
\newcommand{\I}[1]{\mathbb{I}(#1)}
\newcommand{\e}{\varepsilon}
\renewcommand{\epsilon}{\e}
\newcommand{\hR}{\widehat{R}}

\newcommand{\Str}{S_\text{Tr}}
\newcommand{\Ste}{S_\text{Te}}

\newcommand{\Prob}[1]{\mathbb{P}(#1)}

\makeatletter
\renewcommand*{\@fnsymbol}[1]{\ensuremath{\ifcase#1\or *\or \dagger\or \ddagger\or
    \mathsection\or \mathparagraph\or \|\or **\or \dagger\dagger
    \or \ddagger\ddagger \else\@ctrerr\fi}}
\makeatother

\title{Detecting Overfitting via Adversarial Examples}

\author{%
Roman Werpachowski%
\hspace{1cm} Andr\'as Gy\"orgy%
\hspace{1cm} Csaba Szepesv\'ari%
\\
DeepMind, London, UK \\ 
\texttt{\{romanw,agyorgy,szepi\}@google.com} \\
}

\begin{document}
\maketitle

\begin{abstract}
The frequent reuse of test sets in popular benchmark problems raises doubts about the credibility of  reported test-error rates. Verifying whether a learned model is overfitted to a test set is challenging as independent test sets drawn from the same data distribution are usually unavailable, while other test sets may introduce a distribution shift.
We propose a new hypothesis test that uses only the original test data to detect overfitting. 
It utilizes a new unbiased error estimate that is based on adversarial examples generated from the test data and importance weighting. Overfitting is detected if this error estimate is sufficiently different from the original test error rate.
We develop a specialized variant of our test for multiclass image classification, and apply it to testing overfitting of recent models to the popular ImageNet benchmark. Our method correctly indicates overfitting of the trained model to the training set, but is not able to detect any overfitting to the test set, in line with other recent work on this topic.
\vspace{-0.2cm}
\end{abstract}

\bigskip

\setcounter{footnote}{0}

\section{Introduction}

\begin{wrapfigure}[15]{r}{0.5\textwidth}
	\vspace{-0.45cm}
  \begin{center}
    \includegraphics[scale=0.44]{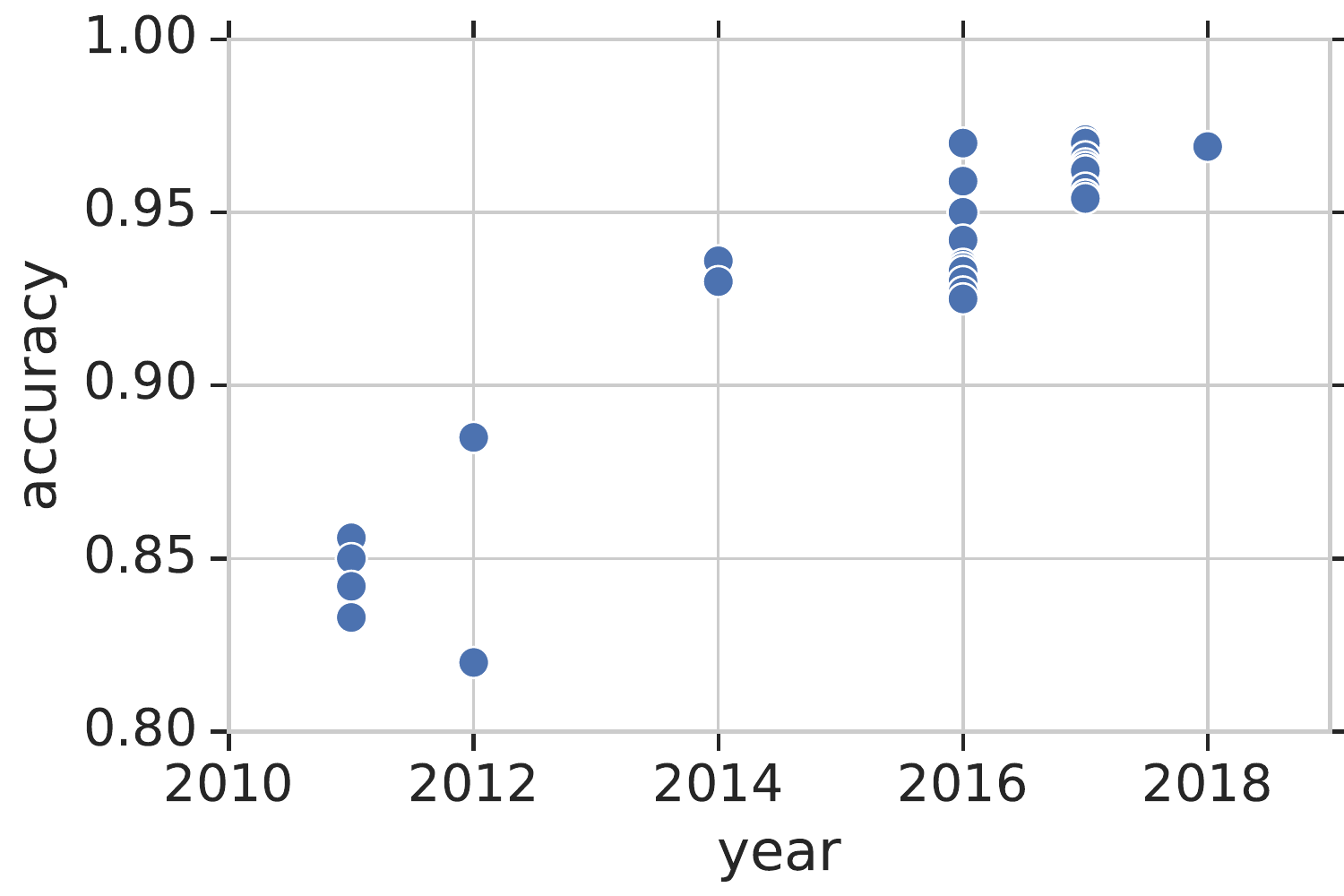}
	\captionsetup{font=small}
    \caption{Accuracy of image classifiers on the CIFAR-10 test set, by year of publication (data from \citep{recht2018cifar10.1}).}
    \label{fig:cifar10-test-accuracy}
    \end{center}
    \vspace{-0.4cm}
\end{wrapfigure}

Deep neural networks achieve impressive performance on many important machine learning benchmarks, such as image classification~\citep{CIFAR10,Krizhevsky2012,Inception,Simonyan15,he2016deep}, automated translation \citep{BahdanauCB14,WuSCLNMKCGMKSJL16} or speech recognition \citep{DengSpeechRecog,graves2013speech}. 
However, the benchmark datasets are used a multitude of times by researchers worldwide. Since state-of-the-art methods are selected and published based on their performance on the corresponding test set, it is typical to see results that continuously improve over time; see, e.g., the discussion of \citet{recht2018cifar10.1} and  \Cref{fig:cifar10-test-accuracy} for the performance improvement of classifiers published for the popular CIFAR-10 image classification benchmark \citep{CIFAR10}. 

This process may naturally lead to models overfitted to the test set, rendering test error rate (the average error measured on the test set) an unreliable indicator of the actual performance.
Detecting whether a model is overfitted to the test set is challenging, since independent test sets drawn from the same data distribution are generally not available, while alternative test sets often introduce a distribution shift.

To estimate the performance of a model on unseen data, one may use generalization bounds to get upper bounds on the expected error rate. 
The generalization bounds are also applicable when the model and the data are dependent (e.g., for cross validation or for error estimates based on the training data or the reused test data), but they usually lead to loose error bounds. 
Therefore, although much tighter bounds are available if the test data and the model are independent, comparing confidence intervals constructed around the training and test error rates leads to an underpowered test for detecting the dependence of a model on the test set.
Recently, several methods have been proposed that allow the reuse of the test set while keeping the validity of test error rates  \citep{Dwork636}. However, these are \emph{intrusive}: they require the user to follow a strict protocol of interacting with the test set and are thus not applicable in the more common situation when enforcing such a protocol is impossible.

\begin{wrapfigure}[15]{r}{0.5\textwidth}
\vspace{-0.5cm}
\begin{center}
\begin{tabular}{ccc}
\includegraphics[width=0.18\textwidth]{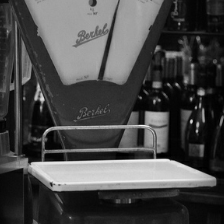} &\mbox{}\hspace{0.1cm} & \includegraphics[width=0.18\textwidth]{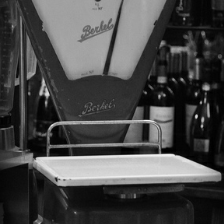} \\
\small scale, weighing machine & & \small toaster
\end{tabular}
\vspace{-0.3cm}
\end{center}
\captionsetup{font=small}
\caption{Adversarial example for the ImageNet dataset and the VGG16 classifier of \citep{Simonyan15}, generated by a $(5, -5)$ translation: the original example (left) is correctly classified as ``scale, weighing machine,'' the adversarially generated example (right) is classified as ``toaster,'' while the image class is the same for any human observer.
\label{fig:adv_image}}
\end{wrapfigure}

In this paper we take a new approach to the challenge of detecting overfitting of a model to the test set,
and devise a \emph{non-intrusive} statistical test that does not restrict the training procedure and is based on the original test data.
To this end, we introduce a new error estimator that is less sensitive to overfitting to the data; our test rejects the independence of the model and the test data if the new error estimate and the original test error rate are too different.
The core novel idea is that the new estimator is based on adversarial examples  \citep{ExplainAdvExampl},  that is, on data points\footnote{Throughout the paper, we use the words ``example'' and ``point'' interchangeably.} 
that are not sampled from the data distribution, but instead are cleverly crafted based on existing data points so that the model errs on them.
Several authors showed that the best models learned for the above-mentioned benchmark problems are highly sensitive to adversarial attacks~\citep{ExplainAdvExampl,TransMLAdvSamp,Uesato,Carlini017a,Carlini017b,papernot2017practical}:
for instance, one can often create adversarial versions of images properly classified by a state-of-the-art model such that the model
will misclassify them, yet the adversarial perturbations are (almost) undetectable for a human observer; see, e.g., \Cref{fig:adv_image}, where the adversarial image is obtained from the original one by
a carefully selected translation.

The \emph{adversarial (error) estimator} proposed in this work uses adversarial examples (generated from the test set) together with importance weighting to take into account the change in the data distribution (covariate shift) due to the adversarial transformation. The estimator is unbiased and has a smaller variance than the standard test error rate if the test set and the model are independent.%
\footnote{Note that the adversarial error estimator's goal is to estimate the error rate, not the adversarial error rate (i.e., the error rate on the adversarial examples).} 
 More importantly, since it is based on adversarially generated data points, the adversarial estimator is expected to differ significantly from the test error rate if the model is overfitted to the test set, providing a way to detect test set overfitting. Thus, the test error rate and the adversarial error estimate (calculated based on the same test set) must be close if the test set and the model are independent, and are expected to be different in the opposite case. In particular, if the gap between the two error estimates is large, the independence hypothesis (i.e., that the model and the test set are independent) is dubious and will be rejected.
Combining results from multiple training runs, we develop another method to test overfitting of a model architecture and training procedure (for simplicity,  throughout the paper we refer to both together as the \emph{model architecture}). %
 The most challenging aspect of our method is to construct adversarial perturbations for which we can calculate importance weights, while keeping enough degrees of freedom in the way the adversarial perturbations are generated to maximize power, the ability of the test to detect dependence when it is present.

To understand the behavior of our tests better, we first use them on a synthetic binary classification problem, where the tests are able to successfully identify the cases where overfitting is present.
Then we apply our independence tests to state-of-the-art classification methods for the popular image classification benchmark,  ImageNet \citep{ImageNet}. As a sanity check, in all cases examined, our test rejects (at confidence levels close to 1) the independence of the individual models from their respective training sets. Applying our method to VGG16 \citep{Simonyan15} and Resnet50 \citep{he2016deep} models/architectures, their \emph{independence to the  ImageNet test set cannot be rejected at any reasonable confidence}. 
This  is in agreement with recent findings of \citep{BenRechtImageNet}, and provides additional evidence that despite of the existing danger, it is likely that no overfitting has happened during the development of ImageNet classifiers.

The rest of the paper is organized as follows: 
In \Cref{sec:formal-model},
we introduce a formal model for error estimation using adversarial examples, including the definition of adversarial example generators. 
The new overfitting-detection tests are derived in \Cref{sec:detection}, and applied to a synthetic problem in \Cref{sec:synthetic}, and to the ImageNet image classification  benchmark in \Cref{sec:numerical-experiments}. Due to space limitations, some auxiliary results, including the in-depth analysis of our method on the synthetic problem, are relegated to the appendix.

\vspace{-0.2cm}
\section{Adversarial Risk Estimation}
\label{sec:formal-model}

We consider a classification problem with deterministic (noise-free) labels, which is a reasonable assumption for many practical problems, such as image recognition (we leave the extension of our method to noisy labels for future work). Let $\X \subset \R^D$ denote the input space and $\Y=\{0,\ldots,K-1\}$ the set of labels.
Data is sampled from the distribution $\cP$ over $\X$, %
and the class label is determined by the \emph{ground truth} function $f^*:\X \to \Y$. %
We denote a random vector drawn from $\cP$ by $X$, and its corresponding class label by $Y=f^*(X)$. We consider deterministic classifiers $f: \X \to \Y$. The performance of $f$ is measured by the zero-one loss: $L(f,x)=\I{f(x) \neq f^*(x)}$,\footnote{For an event $B$, $\I{B}$ denotes its indicator function: $\I{B}=1$ if $B$ happens and $\I{B}=0$ otherwise.} and the \emph{expected error} (also known as the \emph{risk} or \emph{expected risk} in the learning theory literature) of the classifier $f$ is defined as
$R(f) = \E[\I{f(X) \neq Y}] = \int_{\X} L(f, x) \dP(x)$.

Consider a test dataset $S=\{(X_1,Y_1)\ldots,(X_m,Y_m)\}$ where the $X_i$ are drawn from $\cP$ independently of each other and $Y_i=f^*(X_i)$.
In the learning setting, the classifier $f$ usually also depends on some randomly drawn training data, hence is random itself.
If $f$ is (statistically) independent from $S$, then $L(f, X_1),\ldots,L(f,X_m)$ are i.i.d., thus the empirical error rate 
\vspace{-0.1cm}
\begin{equation*}
\hR_S(f) = \frac{1}{m} \sum_{i=1}^m L(f,X_i)= \frac{1}{m} \sum_{i=1}^m \I{f(X_i) \neq Y_i}
\vspace{-0.1cm}
\end{equation*}
is an unbiased estimate of $R(f)$ for all $f$; that is, $R(f)=\E[\hR_S(f)|f]$. If $f$ and $S$ are not independent, the performance guarantees on the empirical estimates available in the independent case are significantly weakened;
for example, in case of overfitting to $S$, the empirical error rate is likely to be much smaller than the expected error. 
Another well-known way to estimate $R(f)$ is to use \emph{importance sampling} (IS) \citep{vKAH49b}:
instead of sampling from the distribution $\cP$, we sample from another distribution $\cP'$ and correct the estimate by appropriate reweighting. Assuming $\cP$ is absolutely continuous with respect to $\cP'$ on the set $E=\{x \in \X: L(f,x)\neq 0 \}$,
$R(f) = \int_{\mathcal{X}} L(f, x) \dP(x) = \int_{E} L(f, x) h(x) \dP'(x)$, %
where $h = \tfrac{\dP}{\dP'}$ is the density (Radon-Nikodym derivative) of $\cP$ with respect to $\cP'$ on $E$ ($h$ can be defined to have arbitrary finite values on $\X\setminus E$).
It is well known that the  the corresponding empirical error estimator
\vspace{-0.2cm}
\begin{align}
\label{eq:hR}
\hR'_{S'}(f)=\frac{1}{m}\sum_{i=1}^m L(f,X'_i) h(X'_i) =\frac{1}{m} \sum_{i=1}^m  \I{f(X'_i) \neq Y'_i} h(X'_i)
\end{align}
obtained from a sample $S'=\{(X'_1,Y'_1),\ldots,(X'_m,Y'_m)\}$ drawn independently from $\cP'$
is unbiased (i.e., $\E[\hR_{S'}(f)|f]=R(f)$) if $f$ and $S'$ are independent. 

The variance of $\hR'_{S'}$
is minimized if $\cP'$ is the so-called zero-variance IS %
distribution, which is supported
on $E$ with $h(x)=\tfrac{R(f)}{L(f,x)}$ for all $x \in E$ (see, e.g., \citep[Section~4.2]{Bucklew04}).
This suggest that an effective sampling distribution $\cP'$ should concentrate on points where $f$ makes mistakes, which also facilitates that 
$\hR'_{S'}(f)$ become large if $f$ is overfitted to $S$ and hence $\hR_S(f)$ is small. We achieve this through the application of adversarial examples.

\subsection{Generating adversarial examples}
\label{sec:aeg}
\vspace{-0.05cm}

\begin{figure}[!h]
\begin{center}
\includegraphics[width=0.6\textwidth]{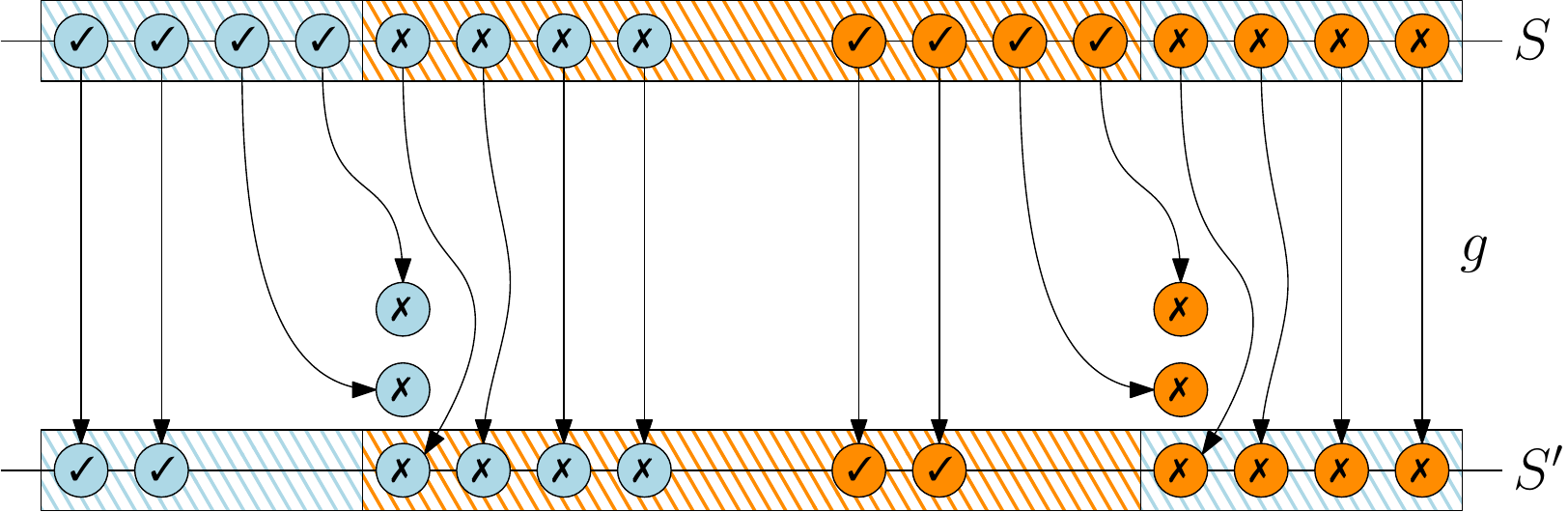}
\end{center}
 \captionsetup{font=small}
 \caption{Generating adversarial examples.  The top row depicts the original dataset $S$, with blue and orange points representing the two classes. The classifier's prediction is represented by the color of the striped areas %
 (checkmarks and crosses denote if a point is correctly or incorrectly classified). The arrows show the adversarial transformations via the AEG $g$, resulting in the new dataset $S'$; misclassified points are unchanged, while some correctly classified points are moved, but their original class label is unchanged. If the original data distribution is uniform over $S$, the transformation $g$ is density preserving, but not measure preserving: after the transformation the two rightmost correctly classified points in each class have probability $0$, while the leftmost misclassified point in each class has probability $3/16$; hence, the density $h_g$ for the latter points is $1/3$.
  \label{fig:AEG}
  }
  \vspace{-0.1cm}
\end{figure}

In this section we introduce a formal framework for generating adversarial examples.
Given a classification problem with data distribution $\cP$ and ground truth $f^*$,
an \emph{adversarial example generator} (AEG) for a classifier $f$ is a (measurable) mapping $g:\X\to\X$ such that
\vspace{-0.1cm}
\begin{enumerate}%
\renewcommand{\theenumi}{G\arabic{enumi}}
\renewcommand{\labelenumi}{(\theenumi)}
\item  \label{g:pres} $g$ preserves the class labels of the samples, that is, $f^*(x)=f^*(g(x))$ for $\cP$-almost all $x$; %
\item  \label{g:correct} $g$ does not change points that are incorrectly classified by $f$, that is, $g(x)=x$ if $f(x) \neq f^*(x)$ for $\cP$-almost all $x$.
\end{enumerate}
\vspace{-0.1cm}

\Cref{fig:AEG} illustrates how an AEG works. %
In the literature, an adversarial example $g(x)$ is usually generated by staying in a small vicinity of the original data point $x$ (with respect to, e.g., the $2$- or the max-norm) and assuming that the resulting label of $g(x)$ is the same as that of $x$ (see, e.g., \citep{ExplainAdvExampl,Carlini017a}). This foundational assumption%
---which is in fact a margin condition on the distribution---is captured in condition~\eqref{g:pres}. %
 \eqref{g:correct} formalizes the fact that there is no need to change samples which are already misclassified. Indeed, existing AEGs comply with this condition.  
 
The performance of an AEG is usually measured by how successfully it generates misclassified examples.
Accordingly, we call a point $g(x)$ a \emph{successful adversarial example} if $x$ is correctly classified by $f$ and $f(g(x))\neq f(x)$ (i.e., $L(f,x)=0$ and $L(f,g(x))=1$).

In the development of our AEGs for image recognition tasks, we will make use of another condition.
For simplicity, we formulate this condition for distributions $\cP$ that have a density $\rho$ with respect to the uniform measure on $\X$, which is assumed to exist
(notable cases are when $\X$ is finite, or $\X=[0,1]^D$ or when $\X = \R^D$; in the latter two cases the uniform measure is the Lebesgue measure).
The assumption states that the AEG needs to be \emph{density-preserving} in the sense that
\begin{enumerate}
\setcounter{enumi}{2}
\renewcommand{\theenumi}{G\arabic{enumi}}
\renewcommand{\labelenumi}{(\theenumi)}
\item \label{g:measure} $\rho(x) = \rho(g(x))$ for $\cP$-almost all $x$. %
\end{enumerate}
Note that a density-preserving map may not be measure-preserving (the latter means that for all measurable $A\subset \X$, $\cP(A) = \cP(g(A))$).

We expect \eqref{g:measure} to hold when $g$ perturbs its input by a small amount and if $\rho$ is sufficiently smooth.
The assumption is reasonable for, e.g., image recognition problems (at least in a relaxed form, $\rho(x) \approx \rho(g(x))$) where we expect that very close images will have a similar likelihood as measured by $\rho$. An AEG employing image translations, which satisfies \eqref{g:measure}, will be introduced in \Cref{sec:numerical-experiments}.
Both \eqref{g:pres} and \eqref{g:measure} can be relaxed (to a soft margin condition or allowing a slight change in $\rho$, resp.) at the price of an extra error term in the analysis that follows.

For a fixed AEG $g:\X \to \X$, let $\cP_g$ denote the distribution of $g(X)$ where $X\sim \cP$ ($\cP_g$ is known as the pushforward measure of $\cP$ under $g$). Further, let $h_g = \frac{d\cP}{d\cP_g}$ on $E = \{x\,:\, L(f,x)\ne 0\}$ and arbitrary otherwise. It is easy to see that  $h_g$ is well-defined on $E$ and $h_g(x) \le 1$ for all $x\in E$: This follows from
the fact that $\cP(A) \le \cP_g(A)$ for any measurable $A\subset E$, which holds since 
\begin{align*}
 \cP_g(A) = \Prob{g(X)\in A}\ge \Prob{g(X)\in A, X\in E} = \Prob{X\in A} = \cP(A),
\end{align*}
where the second to last equality holds because $g(X)=X$ for any $X \in E$ under condition \eqref{g:correct}. 

One may think that  \eqref{g:measure}  implies that $h_g(x)=1$ for all $x\in E$. However, this does not hold.
For example, if $\cP$ is a uniform distribution, any $g:\X \to \supp \cP$ satisfies \eqref{g:measure},
where $\supp \cP \subset \X$ denotes the support of the distribution $\cP$.  This is also illustrated in \Cref{fig:AEG}.

\vspace{-0.05cm}
\subsection{Risk estimation via adversarial examples}
\label{sec:change-of-measure}
\vspace{-0.05cm}

Combining the ideas of this section so far, we now introduce unbiased risk estimates based on adversarial examples.
Our goal is to estimate the error-rate of $f$ through an adversarially generated sample $S'=\{(X_1',Y_1),\ldots,(X_m',Y_m)\}$ obtained through an AEG $g$, where $X_i'=g(X_i)$ with $X_1,\ldots,X_m$ drawn independently from $\cP$ and $Y_i=f^*(X_i)$. Since $g$ satisfies \eqref{g:pres} by definition, the original example $X_i$ and the corresponding adversarial example $X'_i$ have the same label $Y_i$. 
Recalling that $h_g = \dP/\dP_g\le 1$ on $E=\{x\in \X\,:\, L(f,x)=1\}$, one can easily show
that the importance weighted adversarial estimate 
\begin{align}
\label{eq:hR2}
\hR_g(f)=\frac{1}{m} \sum_{i=1}^m  \I{f(X_i') \neq Y_i} h_g(X_i')
\end{align}
obtained from \eqref{eq:hR} for the adversarial sample $S'$ has smaller variance than that of the empirical average $\hR_S(f)$, while both are unbiased estimates of $R(f)$.
Recall that both $\hR_g(f)$ and $\hR_S(f)$ are unbiased estimates of $R(f)$ with expectation $\E[\hR_g(f)]=\E[\hR_S(f)]=R(f)$, and so
\begin{align*}
\Var[\hR_g(f)] &= \frac{1}{m}\left( \E[L(f,g(X))^2 h_g(g(X))^2] - R(f)^2 \right) \\
& \le  \frac{1}{m} \left( \E[L(f,g(X)) h_g(g(X))] - R^2(f) \right) = \frac{1}{m}\left( R(f) - R^2(f) \right) = \Var[\hR_S(f)]~.
\end{align*}
Intuitively, the more successful the AEG is (i.e., the more classification error it induces), the smaller the variance of the estimate $\hR_g(f)$ becomes.

\section{Detecting overfitting}
\label{sec:detection}

In this section we show how the risk estimates introduced in the previous section can be used to test the \emph{independence hypothesis} that
\begin{enumerate}
\renewcommand{\theenumi}{H}
\renewcommand{\labelenumi}{(\theenumi)}
\item  \label{H}
the sample $S$ and the model $f$ are independent.
\end{enumerate}
If \eqref{H} holds, $\E[\hR_g(f)]=\E[\hR_S(f)]=R(f)$, and so the difference $T_{S,g}(f)=\hR_g(f) - \hR_S(f)$ is expected to be small.
On the other hand,  if $f$ is overfitted to the dataset $S$ (in which case $\hR_S(f)<R(f)$), we expect $\hR_S(f)$ and $\hR_g(f)$ to behave differently (the latter being less sensitive to overfitting) since (i) $\hR_g(f)$ depends also on examples previously unseen by the training procedure; (ii) the adversarial transformation $g$ aims to increase the loss, countering the effect of overfitting; (iii) especially in high dimensional settings, in case of overfitting one may expect that there are misclassified points very close to the decision boundary of $f$ which can be found by a carefully designed AEG. Therefore, intuitively, \eqref{H} can be rejected if $|T_{S,g}(f)|$ exceeds some appropriate threshold.

\subsection{Test based on confidence intervals}
\label{sec:basic-test}

The simplest way to determine the threshold is based on constructing confidence intervals for these estimator based on concentration inequalities. Under \eqref{H}, standard concentration inequalities, such as the Chernoff or empirical Bernstein bounds \citep{BoLuMa13}, can be used to quantify how fast $\hR_S$ and $\hR_g(f)$ concentrate around the expected error $R(f)$. In particular, we use the following empirical Bernstein bound \citep{EmpBernStop}: Let $\bar{\sigma}_S^2=(1/m) \sum_{i=1}^m (L(f,X_i)-\hR_S(f))^2$ and $\bar{\sigma}_g^2=(1/m) \sum_{i=1}^m (L(f,g(X_i))h_g(g(X_i))-\hR_g(f))^2$ denote the empirical variance of $L(f,X_i)$ and $L(f,g(X_i))h_g(g(X_i))$, respectively. Then, for any $0<\delta\le1$, with probability at least $1-\delta$, 
\begin{equation}
\label{eq:bernstein1}
|\hR_S(f) - R(f)| \le B(m,\bar{\sigma}^2_S,\delta,1), %
\end{equation}
where
$B(m,\sigma^2,\delta,1)=\sqrt{\frac{2\sigma^2\ln(3/\delta)}{m}}+\frac{3 \ln(3/\delta)}{m}$
and we used the fact that the range of $L(f,x)$ is $1$ (the last parameter of $B$ is the range of the random variables considered).
Similarly, with probability at least $1-\delta$, 
\begin{equation}
\label{eq:bernstein2}
|\hR_g(f) - R(f)| \le B(m,\bar{\sigma}^2_g,\delta,1).
\end{equation}

It follows trivially from the union bound that if the independence hypothesis \eqref{H} holds, the above two confidence intervals $[\hR_S(f) - B(m,\bar{\sigma}^2_S,\delta,1), \hR_S(f)+B(m,\bar{\sigma}^2_S,\delta,1)]$ and
$[\hR_g(f) - B(m,\bar{\sigma}^2_g,\delta,1), \hR_S(f)+B(m,\bar{\sigma}^2_g,\delta,1)]$, which both contain $R(f)$ with probability at least $1-\delta$, intersect with probability at least $1-2\delta$.

On the other hand, if $f$ and $S$ are not independent, the performance guarantees \eqref{eq:bernstein1} and \eqref{eq:bernstein2} may be violated and the confidence intervals may become disjoint. If this is detected, we can reject the independence hypothesis \eqref{H} at a confidence level $1-2\delta$ or, equivalently, with $p$-value $2\delta$. In other words, we reject \eqref{H} if the absolute value of the difference of the estimates $T_{S,g}(f)=\hR_g(f) - \hR_S(f)$ exceeds the threshold $B(m,\bar{\sigma}^2_S,\delta,1)+ B(m,\bar{\sigma}^2_g,\delta,1)$ (note that $\E[T_{S,g}(f)=0]$ if $S$ and $f$ are independent).

\subsection{Pairwise test}
\label{sec:pairwise-test}
 
A smaller threshold for $|T_{S,g}(f)|$, and hence a more effective independence test, can be devised if instead of independently estimating the behavior of $\hR_S$ and $\hR_g(f)$, one utilizes their apparent correlation. Indeed, $T_{S,g}(f) = (1/m)\sum_{i=1}^m T_{i,g}(f)$ where
\begin{align}
  \label{eq:Ti}
  T_{i,g}(f) &= L(f,g(X_i))h_g(g(X_i))-L(f,X_i)
\end{align}
and the two terms in $T_{i,g}(f)$ have the same mean and are typically highly correlated by the construction of $g$. Thus, we can apply the empirical Bernstein bound  \citep{EmpBernStop} to the pairwise differences $T_{i,g}(f)$ to set a tighter threshold in the test: if the independence hypothesis \eqref{H} holds (i.e., $S$ and $f$ are independent), then for any $0<\delta<1$, with probability at least $1-\delta$,
\begin{equation}
\label{eq:pairwise}
|T_{S,g}(f)| \le B(m,\bar{\sigma}_T^2,\delta,U) 
\end{equation}
with
$B(m,\sigma^2,\delta,U)=\sqrt{\frac{2\sigma^2\ln(3/\delta)}{m}}+\frac{3U \ln(3/\delta)}{m}$,
where $\bar{\sigma}_T^2 = (1/m)\sum_{i=1}^m (T_i (f)- T_{S,g}(f))^2$ is the empirical variance of the $T_{i,g}(f)$ terms and $U = \sup T_{i,g}(f) - \inf T_{i,g}(f)$;
we also used the fact that the expectation of each $T_{i,g}(f)$, and hence that of $T_{S,g}(f)$, is zero. Since $h_g \le 1$ if $L(f,x)=1$ (as discussed in \Cref{sec:change-of-measure}), it follows that $U \le 2$, but further assumptions (such as $g$ being density preserving) can result in tighter bounds.

This leads to our pairwise dependence detection method:
\begin{center}
\emph{if $|T_{S,g}(f)| > B(m,\bar{\sigma}_T^2,\delta,2)$, reject \eqref{H} at a confidence level $1-\delta$ ($p$-value $\delta$).}
\end{center}
For a given statistic $(|T_{S,g}(f)|, \bar{\sigma}_T^2)$, the largest confidence level (smallest $p$-value) at which \eqref{H} can be rejected can be calculated by setting the value of the statistic $|T_{S,g}(f)|-B(m,\bar{\sigma}_T^2,\delta,2)$ to zero and solving for $\delta$. This leads to the following formula for the $p$-value (if the solution is larger than $1$, which happens when the bound \eqref{eq:pairwise} is loose, $\delta$ is capped at $1$):
\begin{equation}
\label{eq:delta}
\delta=\min\left\{ 1, 3e^{-\frac{m}{9 U^2}\left(\bar{\sigma}_T^2 + 3U |T_{S,g}(f)| - \bar{\sigma}_T\sqrt{\bar{\sigma}_T^2 + 6 U |T_{S,g}(f)}|\right)}\right\}.
\end{equation}
Note that in order for the test to work well, we not only need the test statistic $T_{S,g}(f)$ to have a small variance in case of independence (this could be achieved if $g$ were the identity), but we also need the estimators $\hR_S(f)$ and $\hR_g(f)$ behave sufficiently differently if the independence assumption is violated. The latter behavior is encouraged by stronger AEGs, as we will show empirically in \Cref{sec:vgg16-imagenet} (see \Cref{fig:imagenet-train-transl} in particular).

\subsection{Dependence detector for randomized training}
\label{sec:detection-randomised}

The dependence between the model and the test set can arise from (i) selecting the ``best'' random seed in order to improve the test set performance and/or (ii) tweaking the model architecture (e.g., neural network structure) and hyperparameters (e.g., learning-rate schedule). If one has access to a single instance of a trained model, these two sources cannot be disentangled. However, if the model architecture and training procedure is fully specified and computational resources are adequate, it is possible to isolate (i) and (ii) by retraining the model multiple times and calculating the $p$-value for every training run separately. Assuming $N$ models, let $f_j, j=1,\ldots,N$ denote the $j$-th trained model and $p_j$ the $p$-value calculated using the pairwise independence test \eqref{eq:pairwise} (i.e., from Eq.~\ref{eq:delta} in \Cref{sec:detection}). We can investigate the degree to which (i) occurs by comparing the $p_j$ values with the corresponding test set error rates $R_S(f_j)$.
To investigate whether (ii) occurs, we can average over the randomness of the training runs. 

For every example $X_i \in S$, consider the average test statistic
$\bar{T}_i = \frac{1}{N}\sum_{j=1}^{N} T_{i,g_j}(f_j)$,
where $T_{i,g_j}(f_j)$ is the statistic \eqref{eq:Ti} calculated for example $X_i$ and model $f_j$ with AEG $g_j$ selected for model $f_j$ (note that AEGs are model-dependent by construction). If, for each $i$ and $j$, the random variables $T_i(f_j)$ are independent, then so are the $\bar{T}_i$ (for all $i$). Hence, we can apply the pairwise dependence detector \eqref{eq:pairwise} with $\bar{T}_i $ instead of $T_i$, using the average $\bar{T}_{S} = (1/m) \sum_{i=1}^m \bar{T}_i$ with empirical variance $\bar{\sigma}^2_{T,N} = (1/m) \sum_{i=1}^m (\bar{T}_i - \bar{T}_{S})^2$, %
giving a single $p$-value $p_N$. If the training runs vary enough in their outcomes, different models $f_j$ err on different data points $X_j$, leading to $\bar{\sigma}^2_{T,N} < \bar{\sigma}_T^2$, and therefore strengthening the power of the dependence detector. 
For brevity, we call this independence test an $N$-model test.

\vspace{-0.1cm}
\section{Synthetic experiments}
\label{sec:synthetic}
\vspace{-0.1cm}

\begin{wrapfigure}[16]{r}{0.5\textwidth}
\centering
 \vspace{-0.3cm}
  \includegraphics[width=0.5\columnwidth]{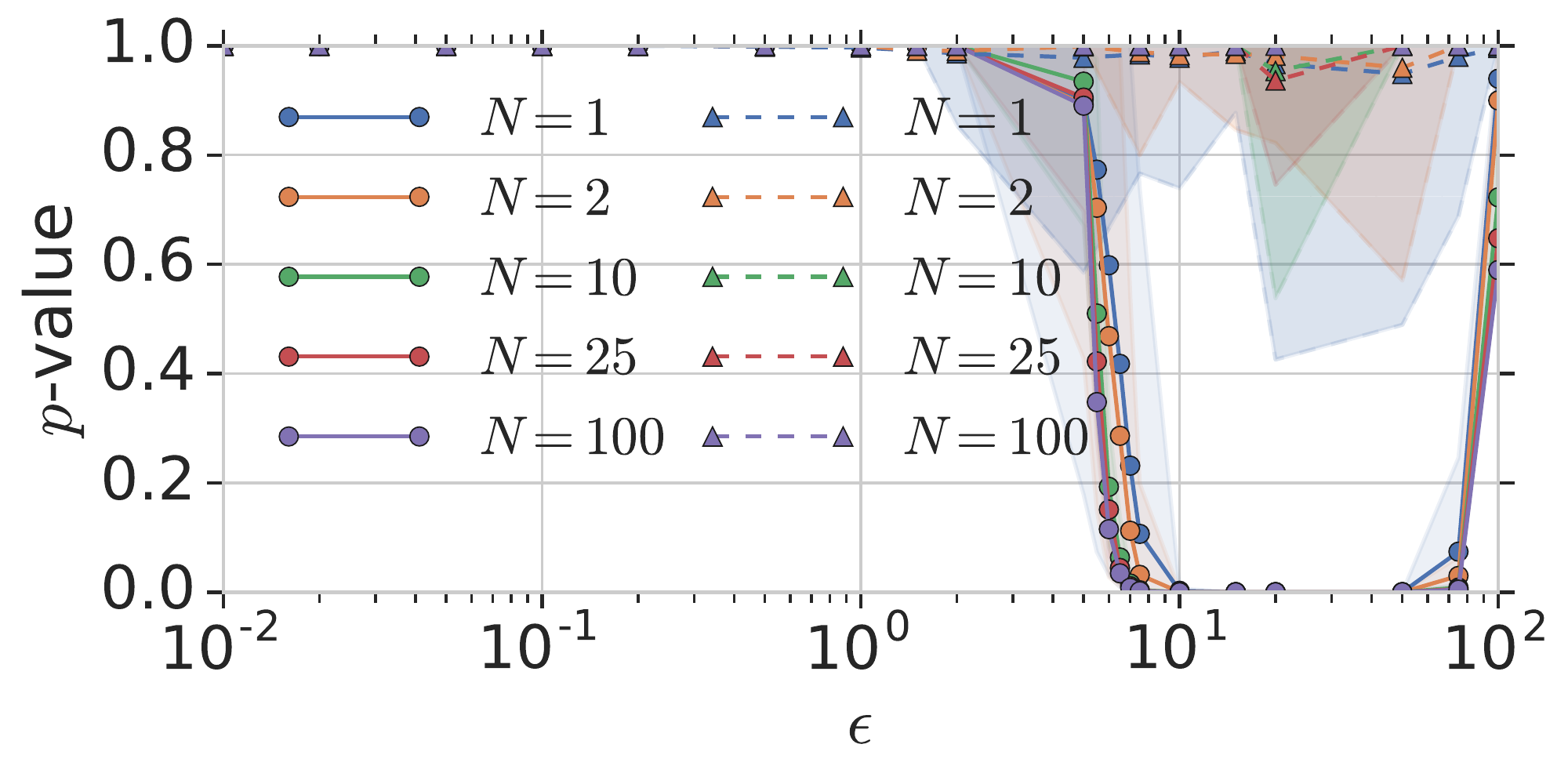}
   \vspace{-0.3cm}
  \captionsetup{font=footnotesize}
  \caption{Average $p$-values produced by the independence test in a separable linear classification problem for the cases of both when the model is independent of (dashed lines) and, resp., dependent on (solid lines) the test set.
 \label{fig:linear-model-s}}
\end{wrapfigure}
First we verify the effectiveness of our method on a simple linear classification problem. Due to space limitations, we only convey high-level results here, details are given in \Cref{app:synthetic}. We assume that the data is linearly separable with a margin and the density $\rho$ is known.
We consider a linear classifiers of the form $f(x) = \sgn(w^\top x + b)$ trained with the cross-entropy loss $c$, and we employ a one-step gradient method (which is an $L_2$ version of the fast gradient-sign method of \cite{ExplainAdvExampl,TransMLAdvSamp}) to define our AEG $g$, which tries to modify a correctly classified point $x$ with label $y$ in the direction of the gradient of the cost function, yielding $x'=x - \epsilon y w/ \| w \|_2$, where $\epsilon \ge 0$ is the strength of the attack. To comply with the requirements for an AEG, we define $g$ as follows:
$g(x)=x'$ if $L(f,x)=0$ and $f^*(x)=f^*(x')$ (corresponding to \eqref{g:correct} and \eqref{g:pres}, respectively), while $g(x)=x$ otherwise. Therefore, if $x'$ is misclassified by $f$, $x$ and $x'$ are the only points mapped to $x'$ by $g$. This simple form of $g$ and the knowledge of $\rho$ allows to compute the density $h_g$, making it easy to compute the adversarial error estimate \eqref{eq:hR2}. \Cref{fig:linear-model-s} shows the average $p$-values produced by our $N$-model independence test for a dependent (solid lines) and an independent (dashed lines) test set. It can be seen that in the dependent case the test can reject independence with high confidence for a large range of attack strength $\epsilon$, while the independence hypothesis is not rejected in the case of true independence. More details (including why only a range of $\epsilon$ is suitable for detecting overfitting) are given in \Cref{app:synthetic}.

\vspace{-0.1cm}
\section{Testing overfitting on ImageNet}
\label{sec:numerical-experiments}
\vspace{-0.1cm}

In the previous section we showed that the proposed adversarial-example-based dependence test works for a synthetic problem where the densities can be computed exactly.
In this section we apply our estimates to a popular image classification benchmark, ImageNet \citep{ImageNet}; here the main issue is to find sufficiently strong AEGs that make computing the corresponding densities possible.

To facilitate the computation of the density $h_g$, we only consider density-preserving AEGs as defined by \eqref{g:measure} (recall that \eqref{g:measure} is different from requiring $h_g = 1$). 
Since in \eqref{eq:hR2} and \eqref{eq:Ti}, $h_g(x)$ is multiplied by $L(f,x)$, we only need to determine the density $h_g$ for data points that are misclassified by $f$.

\subsection{AEGs based on translations}
\label{sec:translation}

To satisfy \eqref{g:measure}, we implement the AEG using translations of images, which have recently been proposed as means of generating adversarial examples \citep{azulay2018}. Although relatively weak, such attacks fit our needs well: unless the images are procedurally centered, it is reasonable to assume that translating them by a few pixels does not change their likelihood.\footnote{Note that this assumption limits the applicability of our method, excluding such centered or essentially centered image classification benchmarks as MNIST \citep{MNIST} or CIFAR-10 \citep{CIFAR10}.}
We also make the natural assumption that the small translations used do not change the true class of an image. Under these assumptions, translations by a few pixels satisfy conditions \eqref{g:pres} and \eqref{g:measure}. An image-translating function $g$ is a valid AEG if it leaves all misclassified images in place (to comply with \eqref{g:correct}), and either leaves a correctly classified image unchanged or applies a small translation.

The main benefit of using  a translational AEG $g$ (with bounded translations) is that its density $h_g(x)$ for an image $x$ can be calculated exactly by considering the set of images $x'$ that can be mapped to $x$ by $g$ (this is due to our assumption \eqref{g:measure}).
We considered multiple ways for constructing translational AEGs. The best version (selected based on initial evaluations on the ImageNet training set), which we called
the \emph{strongest perturbation}, seeks a non-identical neighbor of a correctly classified image $x$ (neighboring images are the ones that are accessible through small translations) that causes the classifier to make an error with the largest confidence.

Formally, we model images as 3D tensors in $[0, 1]^{W \times H \times C}$ space, where $C=3$ for RGB data, and $W$ and $H$ are the width and height of the images, respectively.
Let  $\tau_v(x)$ denote the translation of an image $x$ by $v \in \mathbb{Z}^2$ pixels in the (X, Y) plane (here $\mathbb{Z}$ denotes the set of integers).
To control the amount of change, we limit the magnitude of translations and allow $v \in \V_\e=\{u \in \mathbb{Z}^2: u\neq (0,0), \|u\|_\infty \le \e\}$ only, for some fixed positive $\e$.
Thus, we considers AEGs in the form $g(x) \in \{\tau_v(x): v \in \V\} \cup \{x\}$ if $f(x) = f^*(x)$ and $g(x)=x$ otherwise (if $x$ is correctly classified, we attempt to translate it to find an adversarial example in $\{\tau_v(x): v \in \V\}$ which is misclassified by $f$, but $x$ is left unchanged if no such point exists).
Denoting the density of the pushforward measure $\cP_g$ by $\rho_g$, for any misclassified point $x$,
\vspace{-0.1cm}
\[
\rho_g(x) = \rho(x) + \sum_{v \in \V} \rho(\tau_{-v}(x)) \I{g(\tau_{-v}(x)) = x} = \rho(x)\left(1+ \sum_{v \in \V} \I{g(\tau_{-v}(x)) = x}\right)
\vspace{-0.1cm}
\]
where the second equality follows from \eqref{g:measure}. Therefore, the corresponding density is 
\begin{equation}
\label{eq:hg}
h_g(x) =1/( 1 + n(x) ) \, %
\end{equation}
where $n(x)=\sum_{v \in \V} \I{g(\tau_{-v}(x)) = x}$ is the number of neighboring images which are mapped to $x$ by $g$. Note that given $f$ and $g$, $n(x)$ can be easily calculated by checking all possible translations of $x$ by $-v$ for $v \in \V$.
It is easy to extend the above to non-deterministic perturbations, defined as distributions over AEGs, by replacing the indicator with its expectation $\P(g(\tau_{-v}(x)) = x | x,v)$ with respect to the randomness of $g$, yielding
\begin{equation}
\label{eq:hgnd}
h_g(x) = \frac{1}{1 + \sum_{v \in \V} \P(g(\tau_{-v}(x)) = x | x,v)}~.
\end{equation}
If $g$ is deterministic, we have $h_g(x) \le 1/2$ for any successful adversarial example $x$. Hence, for such $g$,  the range $U$ of the random variables $T_i$ defined in \eqref{eq:Ti} has a tighter upper bound of 3/2 instead 2 (as $T_i \in [-1,1/2]$), leading to a tighter bound in \eqref{eq:pairwise} and a stronger pairwise independence test.
In the experiments, we use this stronger test. We provide additional details about the translational AEGs used in \Cref{app:aeg-image-transl}.

\subsection{Tests of ImageNet models}
\label{sec:vgg16-imagenet}
\vspace{-0.05cm}

We applied our test to check if state-of-the-art classifiers for the ImageNet dataset \citep{ImageNet} have been overfitted to the test set.
In particular, we use the VGG16 classifier of \citep{Simonyan15} and the Resnet50 classifier of \citep{he2016deep}. Due to computational considerations, we only analyzed a single trained VGG16 model, while the Resnet50 model was retrained 120 times. The models were trained using the parameters recommended by their respective authors.

The preprocessing procedure of both architectures involves rescaling every image so that the smaller of width and height is 256 and next cropping centrally to size $224\times 224$. This means that translating the image by $v$ can be trivially implemented by shifting the cropping window by $-v$ without any loss of information for $\| v \|_\infty \le 16$, because we have enough extra pixels outside the original, centrally located cropping window. 
This implies that we can compute the densities of the translational AEGs for any $\| v \|_\infty \le \epsilon = \floor{16/3} = 5$ (see \Cref{app:maxtranslation} for detailed explanation).
Because the ImageNet data collection procedure did not impose any strict requirements on centering the images \citep{ImageNet},  it is reasonable to assume (as we do) that small (lossless) translations respect the density-preserving condition \eqref{g:measure}.

In our first experiment, we applied our pairwise independence test \eqref{eq:pairwise}  with the AEGs described in \Cref{app:aeg-image-transl} (strongest, nearest, and the two random baselines) to all 1,271,167 training examples, as well as to a number of its randomly selected (uniformly without replacement) subsets of different sizes. Besides this being a sanity check, we also used this experiment to select from different AEGs and compare the performance of the pairwise independence test \eqref{eq:pairwise} to the basic version of the test described in \Cref{sec:basic-test}.

\begin{figure}[tbp]
  \centering\begin{tabular}{ccc}
  \includegraphics[width=0.4\textwidth]{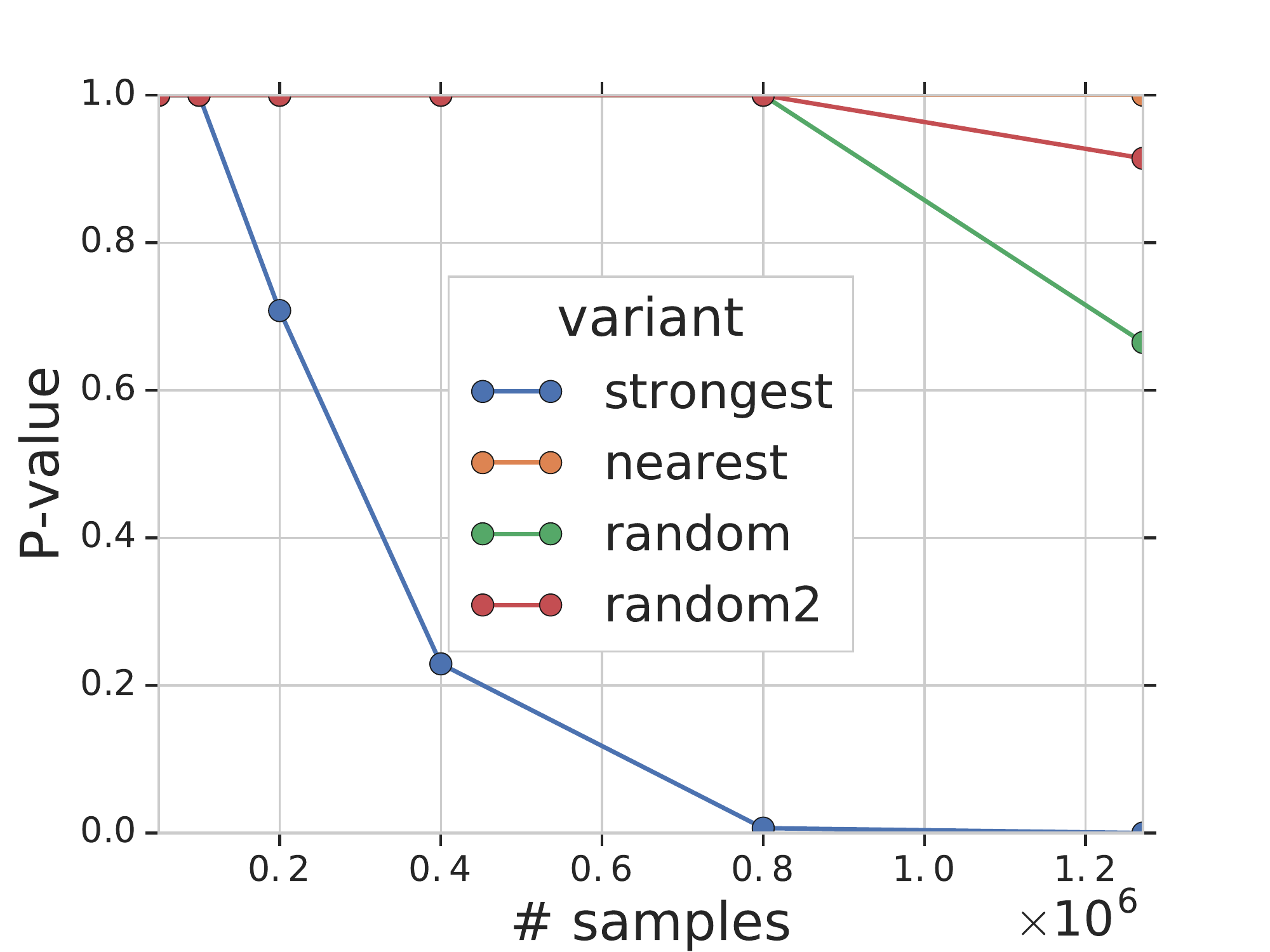} & \mbox{\hspace{0.8cm}} &
  \includegraphics[width=0.4\textwidth]{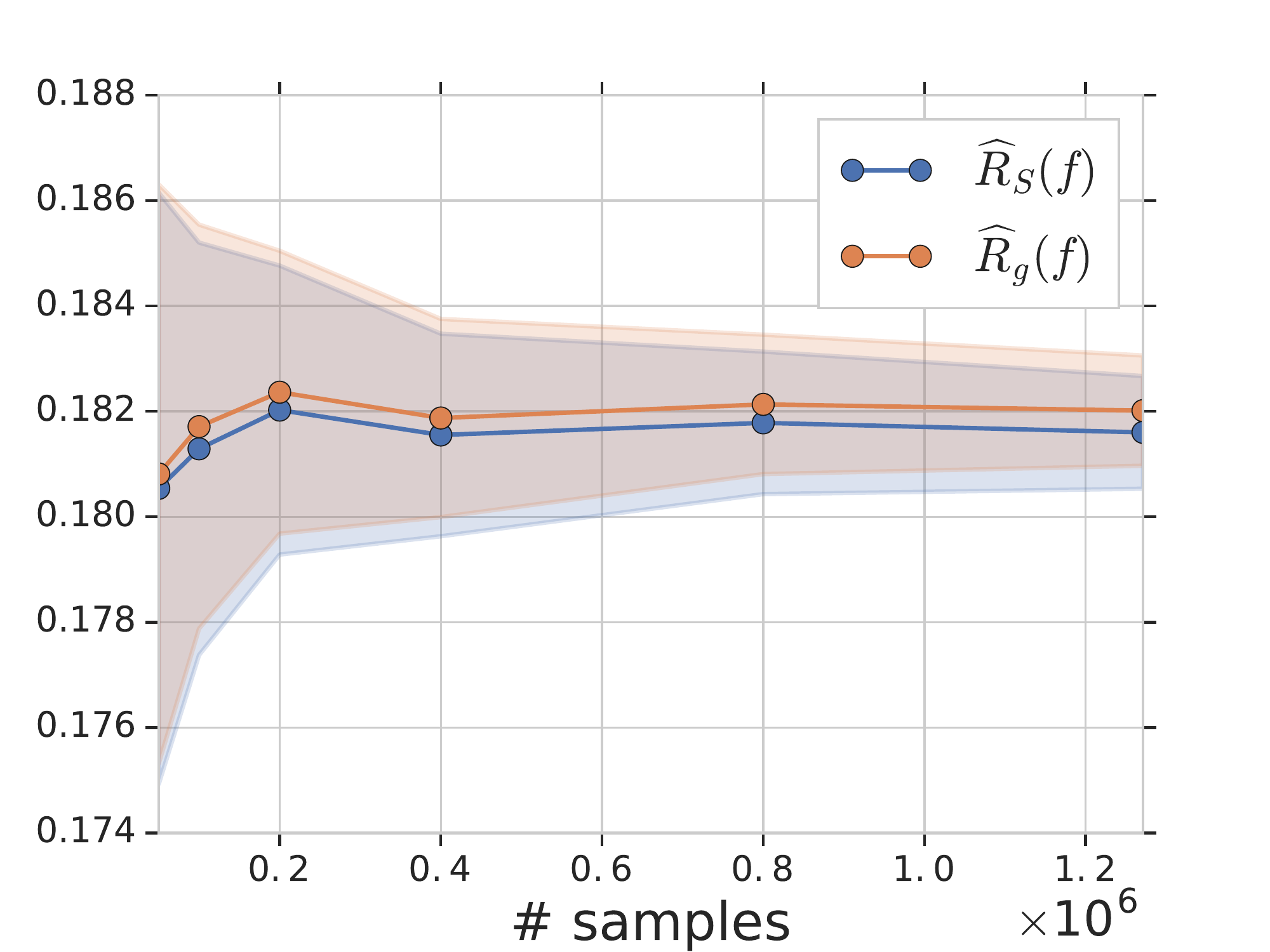}
  \end{tabular}
  \captionsetup{font=small}
  \caption{$p$-values for the independence test on the ImageNet training set for different sample sizes and AEG variants (left); original and adversarial risk estimates, $\hR_S(f)$ and $\hR_g(f)$, on the ImageNet training set with 97.5\% two-sided confidence intervals for the `strongest attack' AEG (right).}
  \label{fig:imagenet-train-transl}
  \vspace{-0.3cm}
\end{figure}

The left graph in \Cref{fig:imagenet-train-transl} shows that with the ``strongest perturbation'', we were able to reject independence of the trained model and the training samples at a confidence level very close to 1 when enough training samples are considered (to be precise, for the whole training set the confidence level is $99.9994\%$). Note, however, that the much weaker ``smallest perturbation'' AEG, as well as the random transformations, are not able to detect the presence of overfitting.
At the same time, the graph on the right hand side shows the relative strength of the pairwise independence test compared to the basic version based on independent confidence interval estimates as described in detail in \Cref{sec:basic-test}: the $97.5\%$-confidence intervals of the error estimates $\hR_S(f)$ and $\hR_g(f)$ overlap, not allowing to reject independence at a confidence level of $95\%$ (note that here $S$ denotes the training set). 

On the other hand, when applied to the test set, we obtained a $p$-value of 0.96, not allowing at all to reject the independence of the trained model and the test set.
This result could be explained by the test being too weak, as no overfitting is detected to the \emph{training} set at similar sample sizes (see \Cref{fig:imagenet-train-transl}), or simply the lack of overfitting. 
Similar results were obtained for Resnet50, where even the $N$-model test with $N=120$ independently trained models resulted a $p$ value of 1, not allowing to reject 
independence at any confidence level.
The view of no overfitting can be backed up in at least two ways:
 first, ``manual'' overfitting to the relatively \emph{large} %
 ImageNet test set is hard. Second, since training an ImageNet model was just too computationally expensive until quite recently, only a relatively small number of different architectures were developed for this problem, and the evolution of their design was often driven by computational efficiency on the available hardware. On the other hand, it is also possible that increasing $N$ sufficiently might show evidence of overfitting (this is left for future work).

\vspace{-0.1cm}
\section{Conclusions}
\label{sec:conclusions}
\vspace{-0.1cm}

We presented a method for detecting overfitting of models to datasets. It relies on an importance-weighted risk estimate from a new dataset obtained by generating adversarial examples from the original data points. 
We applied our method to the popular ImageNet image classification task. For this purpose, we developed a specialized variant of our method for image classification that uses adversarial translations, providing arguments for its correctness.
Luckily, and in agreement with other recent work on this topic \citep{recht2018cifar10.1,BenRechtImageNet,feldman2019multiclass,mania2019modelSimilarity,ChhaviBottou2019ColdCase}, we found no evidence of overfitting of state-of-the-art classifiers to the ImageNet test set.

The most challenging aspect of our methods is to construct adversarial perturbations for which we can calculate the importance weights; finding stronger perturbations than the ones based on translations for image classification is an important question for the future. Another interesting research direction is to consider extensions beyond image classification, for example, by building on recent adversarial attacks for speech-to-text methods \citep{CaWa18}, machine translation \citep{EbrahimiLD18} or text classification \citep{EbrahimiRLD18b}.

\section*{Acknowledgements}

We thank J. Uesato for useful discussions and advice about adversarial attack methods and sharing their implementations \citep{Uesato} with us, as well as M. Rosca and S. Gowal for help with retraining image classification models.
We also thank B. O'Donoghue for useful remarks about the manuscript, and L. Schmidt for an in-depth discussion of their results on this topic.
Finally, we thank D. Balduzzi, S. Legg, K. Kavukcuoglu and J. Martens for encouragement, support, lively discussions and feedback.

{\small
\bibliography{citations}
\bibliographystyle{plainnat}
}

\newpage
\appendix

\section{Synthetic experiments}
\label{app:synthetic}

\begin{figure}[!b]
  \begin{center}
  \vspace{0.6cm}
    \begin{overpic}[width=0.4\textwidth]{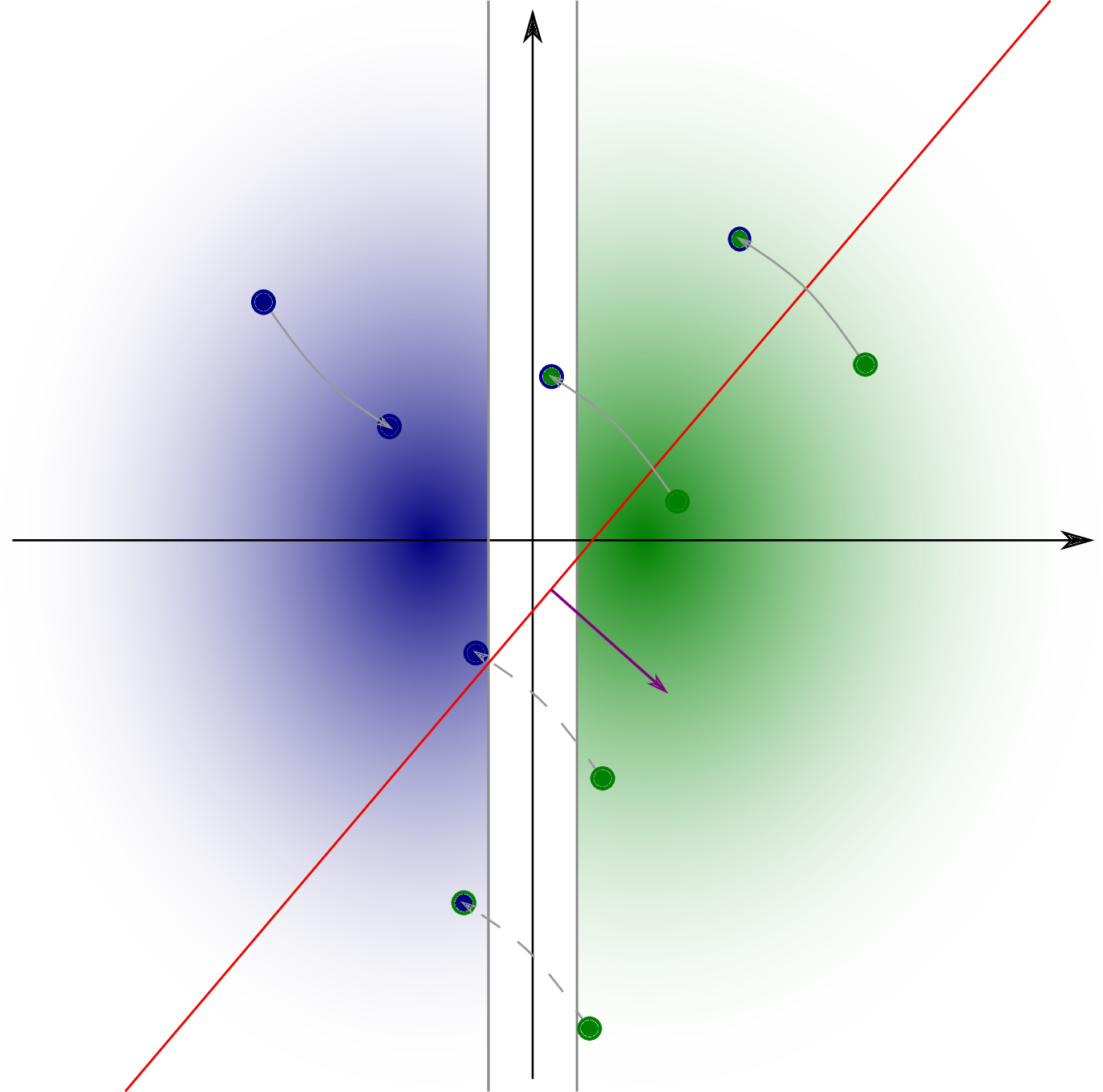}
      \put(49, 51){0}
      \put(95, 52){$x_1$}
      \put(50, 95){$x_2$}
      \put(23, 87){\large$y=-1$}
      \put(55, 87){\large$y=+1$}
      \put(80.5, 65){$x_A$}
      \put(59, 77){$x'_A$}
      \put(63.5, 52.5){$x_B$}
      \put(49.5, 67.5){$x'_B$}
      \put(16, 72){$x_C$}
      \put(33.5, 56){$x'_C$}
      \put(56.5, 27.3){$x_D$}
      \put(35, 39.3){$x'_D$}
      \put(55.5, 5){$x_E$}
      \put(34, 17){$x'_E$}
      \put(55, 41){$w$}
    \end{overpic}
  \end{center}
  \captionsetup{font=small}
  \caption{Illustration of the data distribution and the linear model $f(x) = \sgn(w^\top x + b)$ in two dimensions. The blue and green gradients show the probability density $\rho$ of the data with true labels $y=-1$ and $y=1$, respectively, while the white space between them is the margin with $\rho = 0$. The red line is the model's classification boundary with its parameter vector $w$ shown by the purple arrow. Depending on the label, $w$ or $-w$ is the direction of translation used to perturbed the correctly classified data points, and the translations used by the AEG $g$ for specific points are depicted by grey arrows: 
solid arrows indicate the cases where $g(x) = x' \neq x$, while dashed arrows are for candidate translations which are not performed by the AEG because they would change the true label, $f^*(x') \neq f^*(x)$, and hence $g(x) = x \neq x'$. Each original/perturbed data point is represented by a color-coded circle: the inner color corresponds to the true label (dark blue for $y=-1$ and dark green for $y=1$) while the outer color to the model's prediction (dark blue for $f(x)=-1$ and dark green for $f(x)=1$). Points $x'_A$ and $x'_B$ can be obtained from $x_A$ and $x_B$, respectively, by applying the AEG, $x'_A = g(x_A)$ and $x'_B = g(x_B)$. Since only $x_A$ is mapped to $x'_A$ by $g$, and $g(x'_A) = x'_A$, the density (Radon-Nikodym derivative) can be obtained as $h_g(x'_A) = \rho(x'_A) / (\rho(x_A) + \rho(x'_A)) \in (0, 1)$. In the case of $x'_B$, $h_g(x'_B) = \rho(x'_B) / (\rho(x_B) + \rho(x'_B)) = 0$ due to the margin. Note that the formula for $h_g(x'_A)$ does not depend on whether $x_A$ or $x'_A$ is in the original test set $S$;  in the first case we call $x'_A$ a ``successful adversarial example'' while in the second case $x'_A$ is called ``originally misclassified'' (a similar argument holds for $h_g(x'_B)$). $x'_C$ is not a successful adversarial example since $L(f, x'_C) = 0$ (however, $g(x_C) = x'_C$ according to our definition). Points $x_D$ and $x_E$ are not perturbed by our AEG, since $f^*(x_D) \neq f^*(x'_D)$ and $f^*(x_E) \neq f^*(x'_E)$.
  }
  \label{fig:synthetic-dataset}
\end{figure}
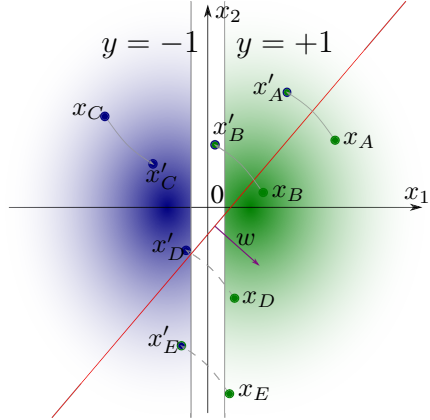

In this section, we present full details of the experiments on a simple synthetic classification problem, which we presented briefly in \Cref{sec:synthetic}. These experiments illustrate the power of the method of \Cref{sec:detection}. The advantage of the simple setup considered here is that we are able to compute the density $h_g$ in an analytic form (see \Cref{fig:synthetic-dataset} for an illustration).

\subsection{Data distribution and model}

Let $\X =\R^{500}$ and consider an input distribution with a density $\rho$ that is an equally weighted mixture of two 500-dimensional isotropic truncated Gaussian distributions $N^\text{trunc}_\pm(\mu_\pm, \sigma^2 I)$ with coordinate-wise standard deviation $\sigma=\sqrt{500}$ ($I$ denotes the identity matrix of size $500\times 500$), means $\mu_\pm = [\pm 1, 0, 0, \ldots, 0]$ and densities $\rho_\pm$ truncated in the first dimension such that $\rho_+(x) = 0$ if $x_1 \le 0.025$ and $\rho_-(x) = 0$ if $x_1 \ge -0.025$. The label of an input point $x$ is $f^*(x)=\sgn(x_1)$, which is the sign of its first coordinate.

We consider linear classifiers of the form $f(x) = \sgn(w^\top x + b)$ trained with the cross-entropy loss $c((w,b),x,y)= \ln (1+ e^{-y(w^\top x +b)})$ where $y=f^*(x)$. We employ a one-step gradient method (which is an $L_2$ version of the fast gradient-sign method of \citep{ExplainAdvExampl,TransMLAdvSamp}) to define our AEG $g$, which tries to modify a correctly classified point $x$ with label $y$ in the direction of the gradient of the cost function $c$: $x'=x + \epsilon \nabla_{x} c((w,b),x,y) / \| \nabla_{x} c((w,b),x,y) \|_2$ for some $\epsilon>0$. For our specific choice of $c$, the above simplifies to $x'=x - \epsilon y w/ \| w \|_2$. To comply with the requirements for an AEG, we define $g$ as follows:
$g(x)=x'$ if $L(f,x)=0$ and $f^*(x)=f^*(x')$ (corresponding to \eqref{g:correct} and \eqref{g:pres}, respectively), while $g(x)=x$ otherwise. Therefore, if $x'$ is misclassified by $f$, $x$ and $x'$ are the only points mapped to $x'$ by $g$. Thus, the density at $x'$ after the transformation $g$ is $\rho'(x')=\rho(x) + \rho(x')(1-L(f,x)) \I{f^*(x)=f^*(x')}$ and
\[
h_g(x') = \frac{\rho(x')}{\rho'(x')}=\frac{\rho(x')}{ \rho(x') + \rho(x) (1-L(f,x)) \I{f^*(x)=f^*(x')}}
\]
(note that $\I{L(f,x)=0}=1-L(f,x)$).

\subsection{Experiment setup}

We present two experiments showing the behavior of our independence test: one where the training and test sets are independent, and another where they are not.

In the first experiment a linear classifier was trained on a training set $\Str$ of size 500 for 50,000 steps using the RMSProp optimizer \citep{Tieleman2012} with batch size 100 and  learning rate 0.01, obtaining zero (up to numerical precision) final training loss $c$ and, consequently, 100\% prediction accuracy on the training data. Then the trained classifier was tested on a large test set $\Ste$ of size 10,000.\footnote{The large number of test examples ensures that the random error in the empirical error estimate is negligible.} Both sets were drawn independently from $\rho$ defined above. We used a range of $\epsilon$ values matched to the scale of the data distribution: from $10^{-2}$, which is the order of magnitude of the margin between two classes (0.05), to $10^2$, which is the order of magnitude of the width of the Gaussian distribution used for each classes ($\sigma=\sqrt{500}$).

In the second experiment we consider the situation where the training and test sets are not independent. To enhance the effects of this dependence, the setup was modified to make the training process more amenable to overfitting by simulating a situation when the model has a wrong bias (this may happen in practice if a wrong architecture or data preprocessing method is chosen, which, despite the modeler's best intentions, worsens the performance). Specifically, during training we added a penalty term $10^4 w_1^2$ to the training loss $c$, decreased the size of the test set to 1000 and used 50\% of the test data for training (the final penalized training loss was 0.25 with 100\% prediction accuracy on the training set). Note that the small training set and the large penalty on $w_1$ yield classifiers that are essentially independent of the only interesting feature $x_1$ (recall that the true label of a point $x$ is $\sgn(x_1)$) and overfit to the noise in the data, resulting in a true model risk $R(f) \approx 1/2$.

\subsection{Results}

\begin{figure}[tbp]
  \vspace{-0.3cm}\centering\begin{tabular}{ccc}
  \vspace{-0.22cm}\includegraphics[width=0.42\textwidth]{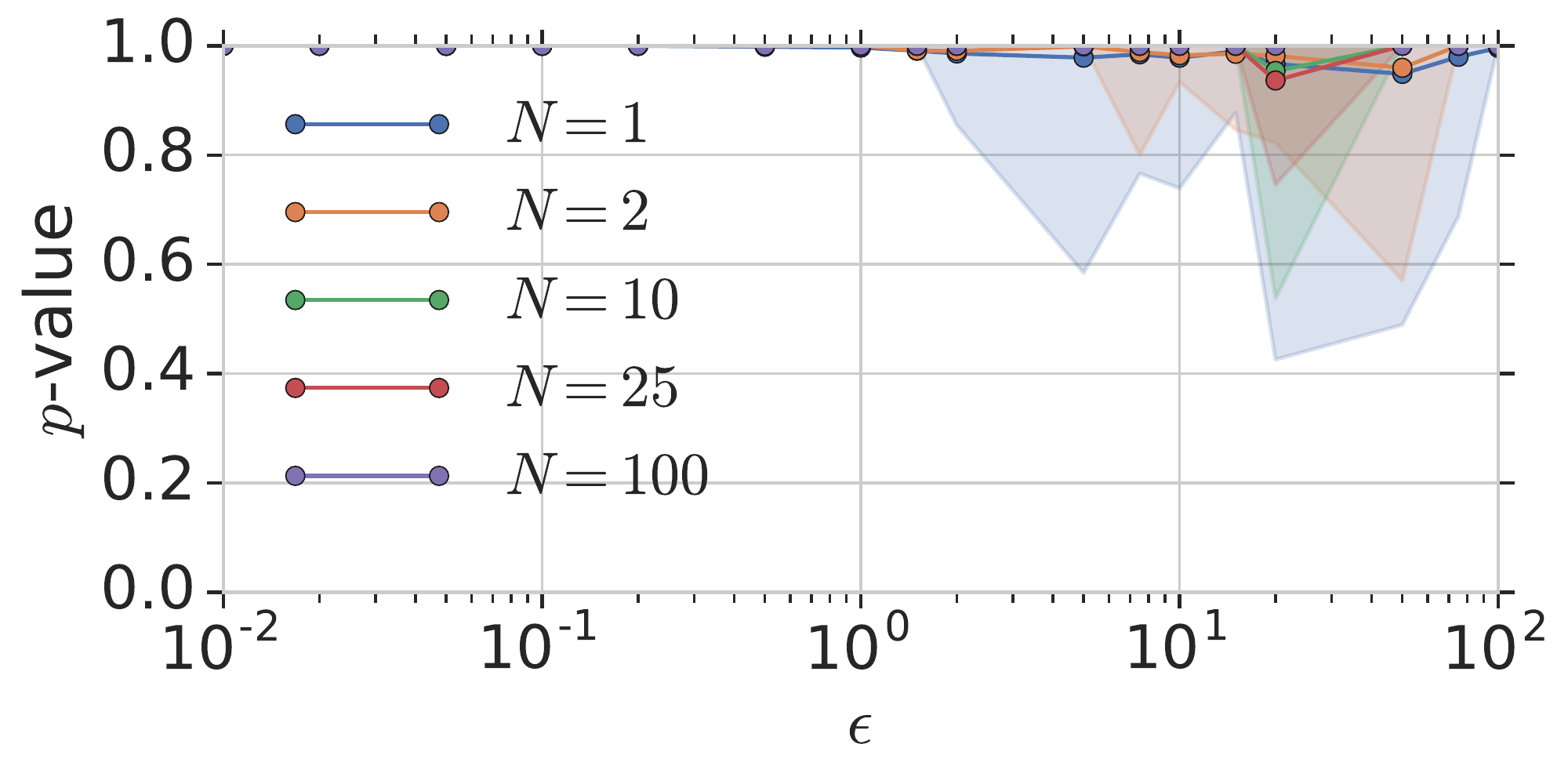} & \mbox{\hspace{0.8cm}} &
  \includegraphics[width=0.42\textwidth]{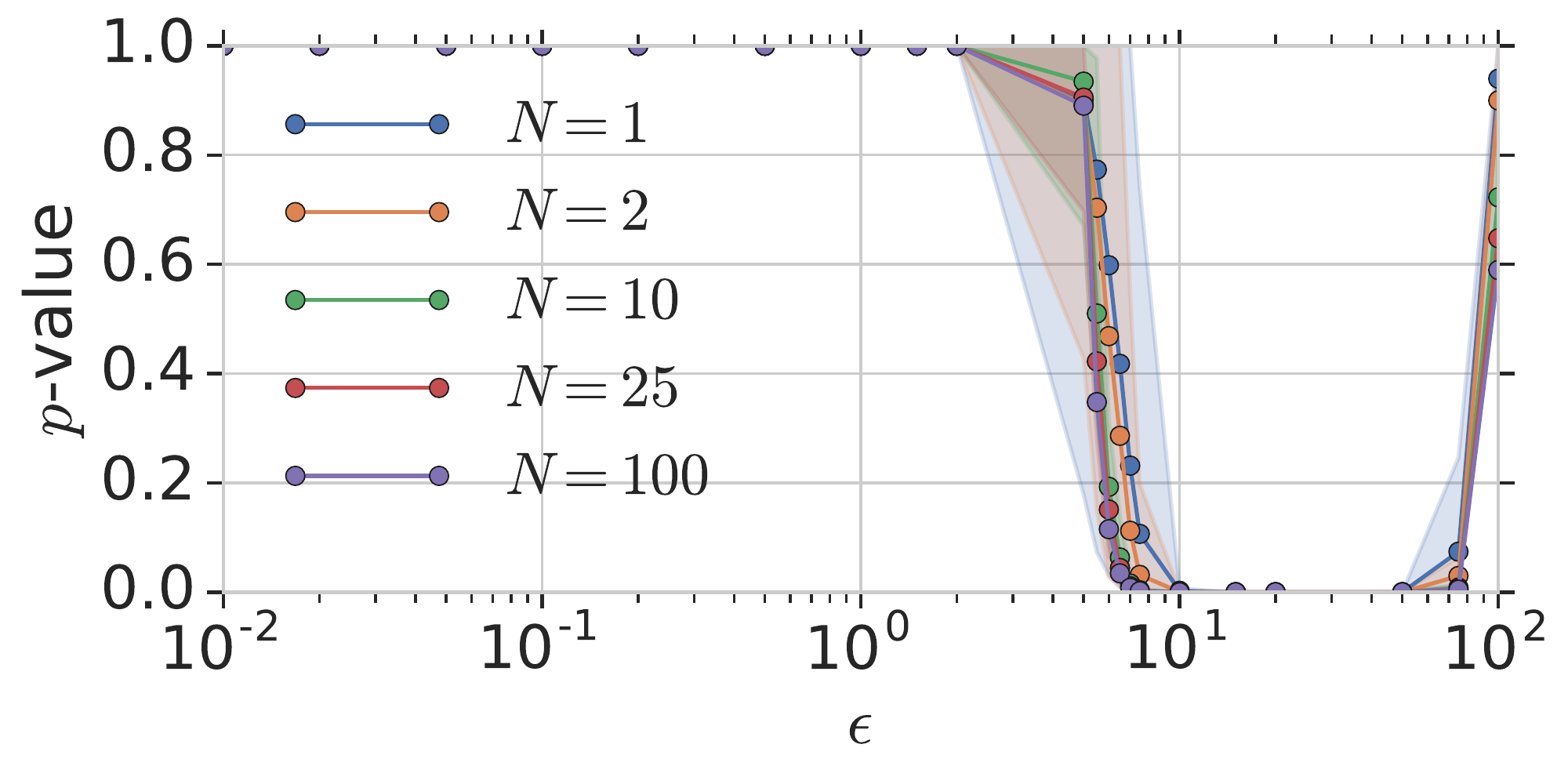} \\
  \vspace{-0.22cm}\includegraphics[width=0.42\textwidth]{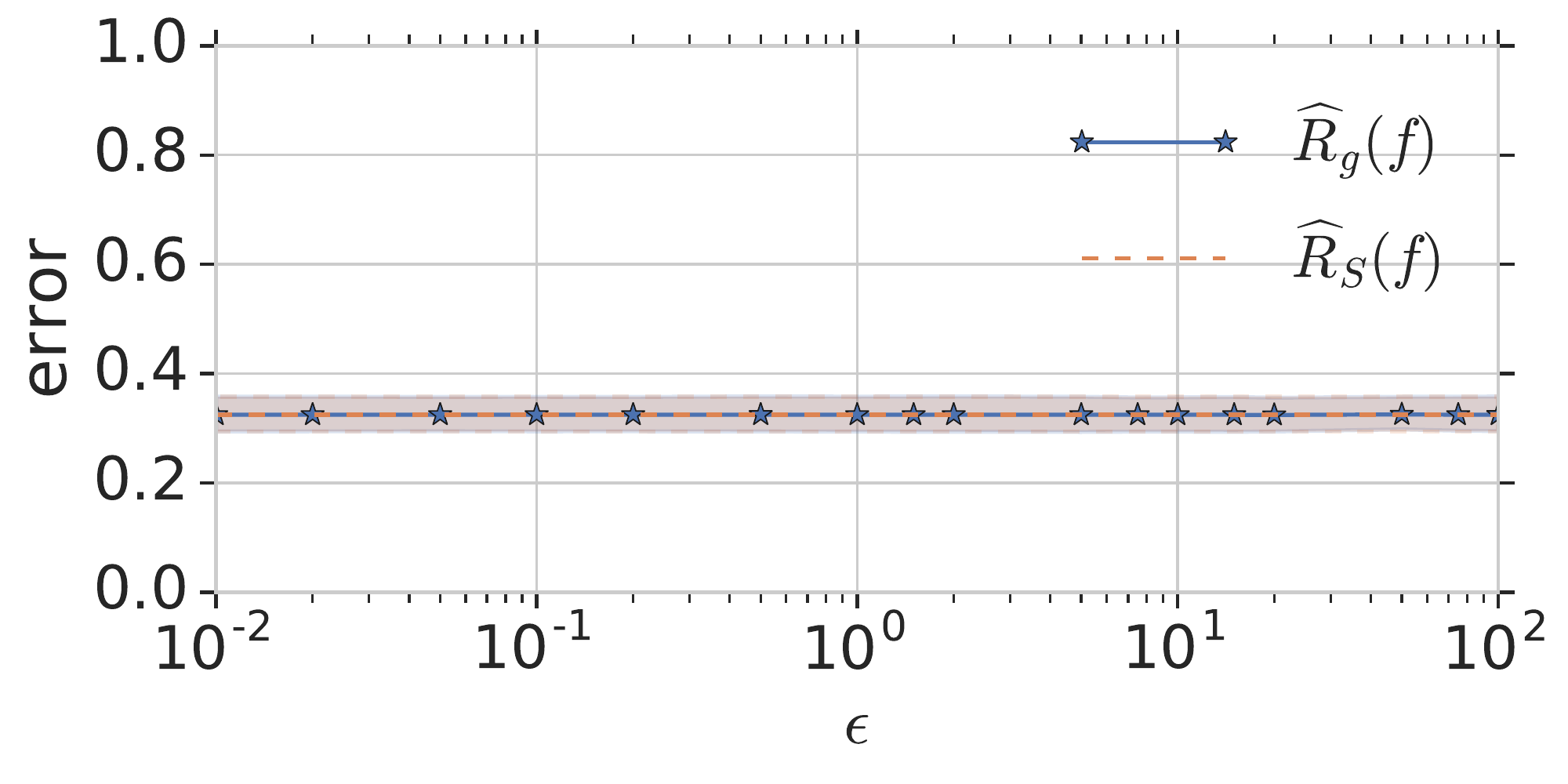} & \mbox{} &
  \includegraphics[width=0.42\textwidth]{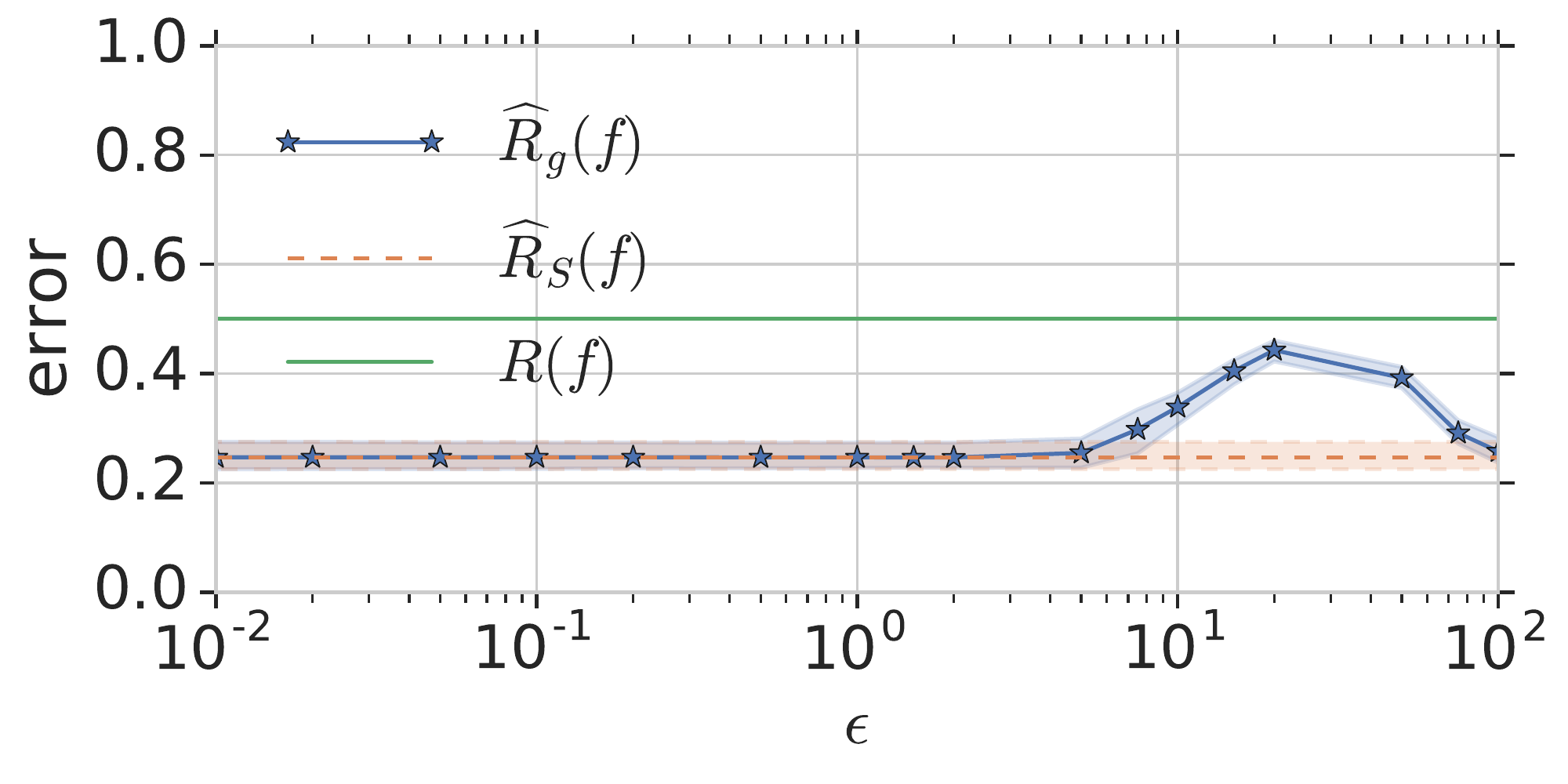} \\
  \vspace{-0.22cm}\includegraphics[width=0.42\textwidth]{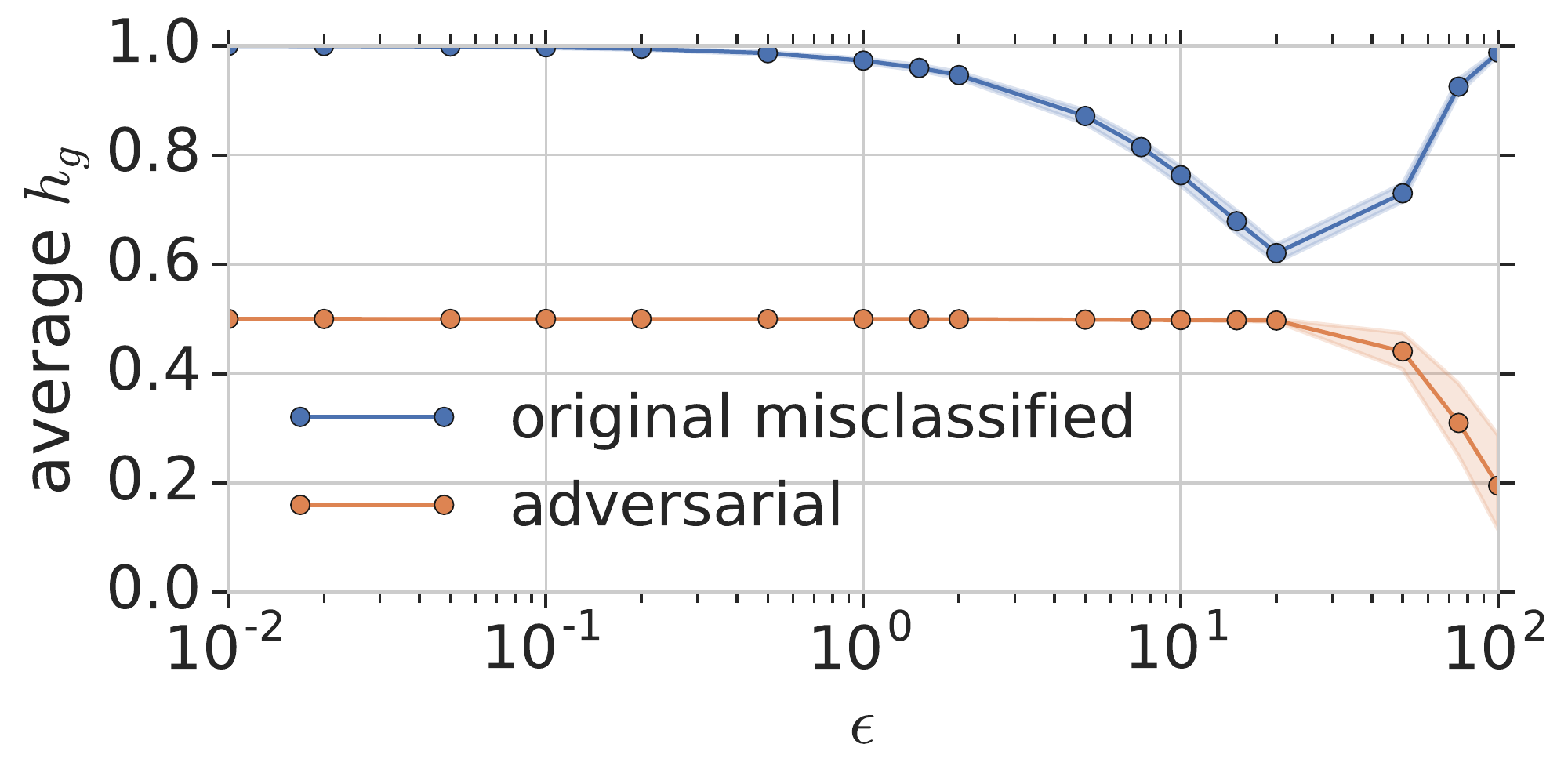} & \mbox{} &
  \includegraphics[width=0.42\textwidth]{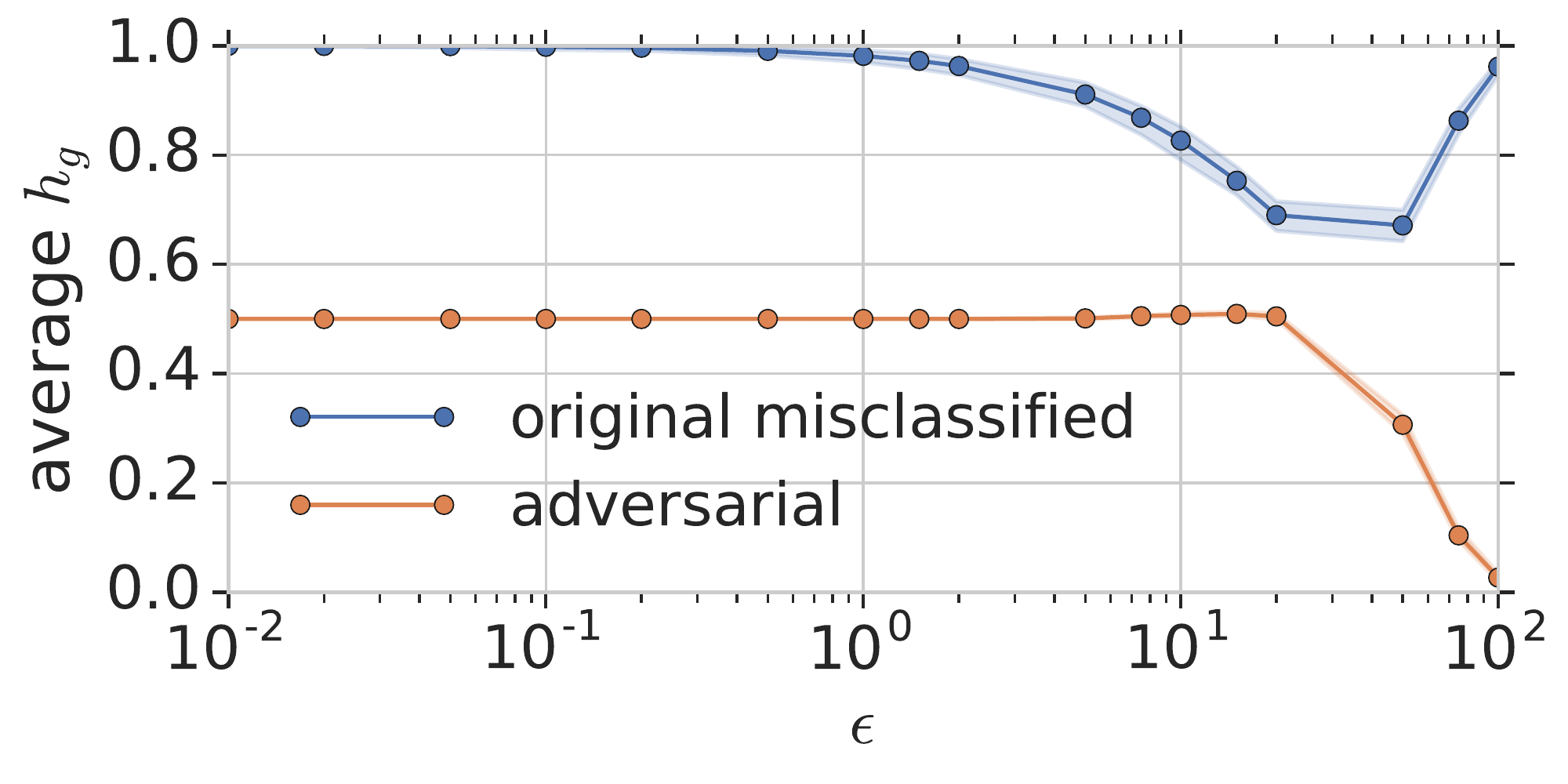} \\  
  \vspace{-0.22cm}\includegraphics[width=0.42\textwidth]{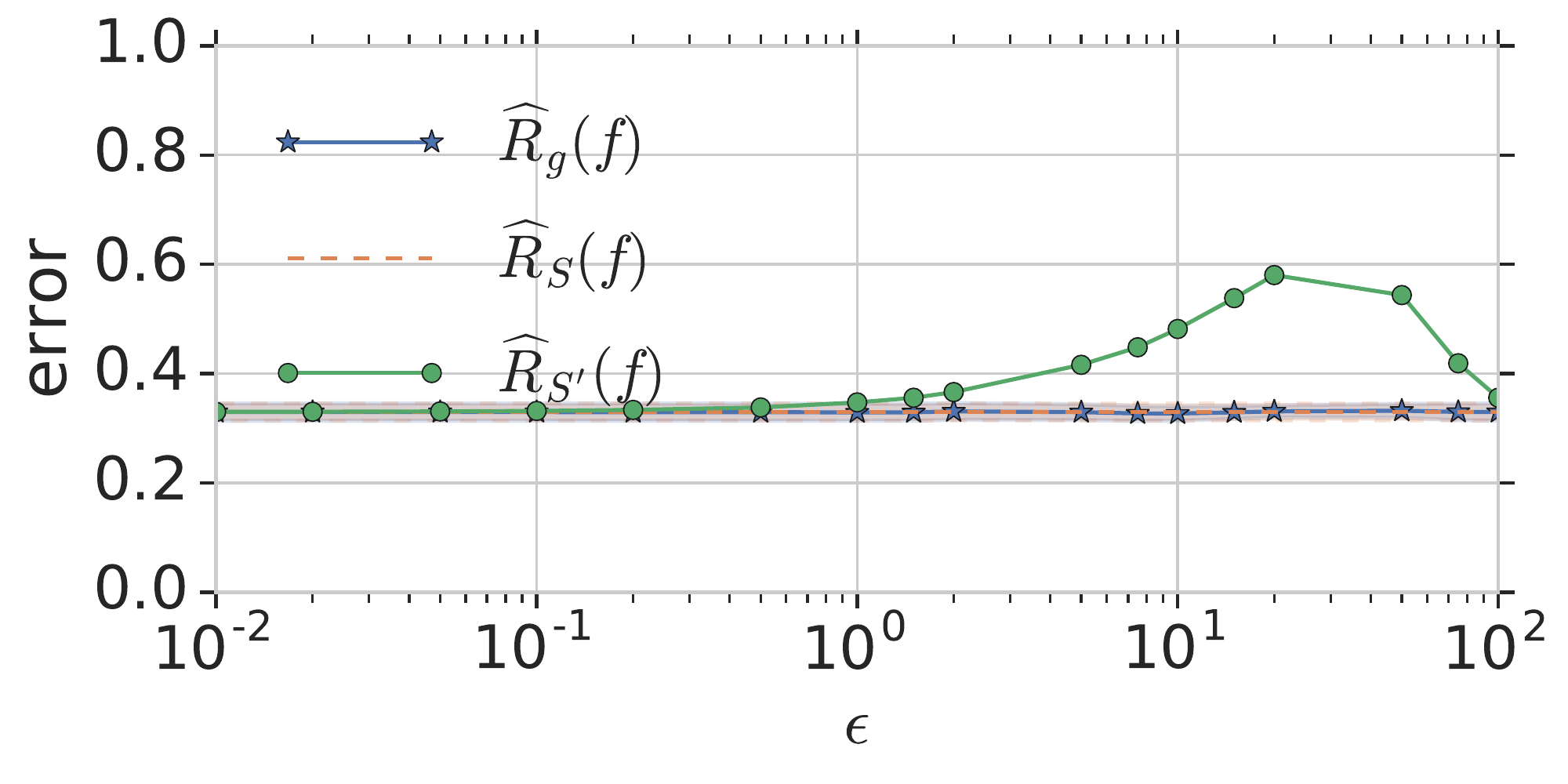} & \mbox{} &
  \includegraphics[width=0.42\textwidth]{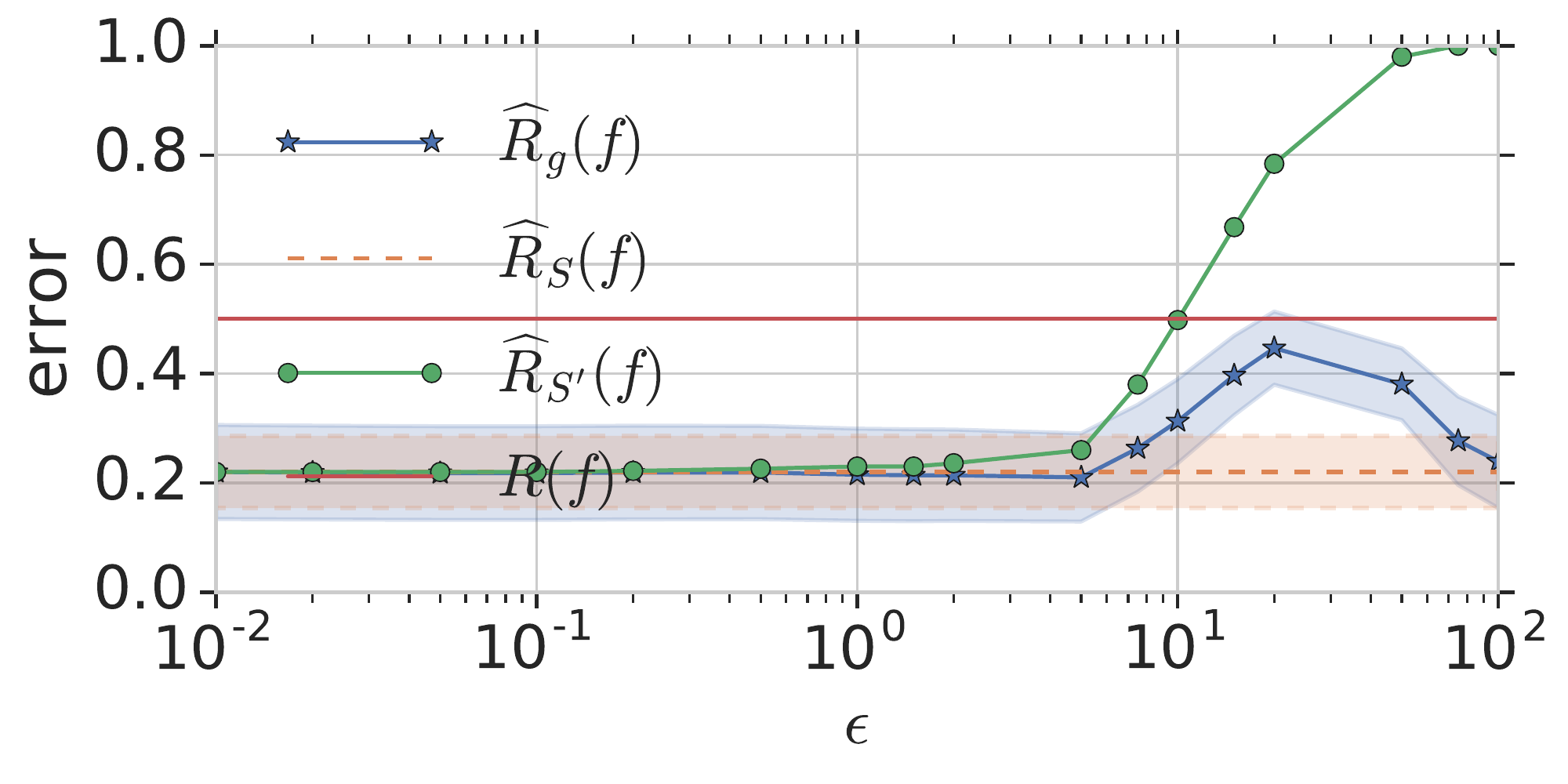} \\
  \vspace{-0.22cm}\includegraphics[width=0.42\textwidth]{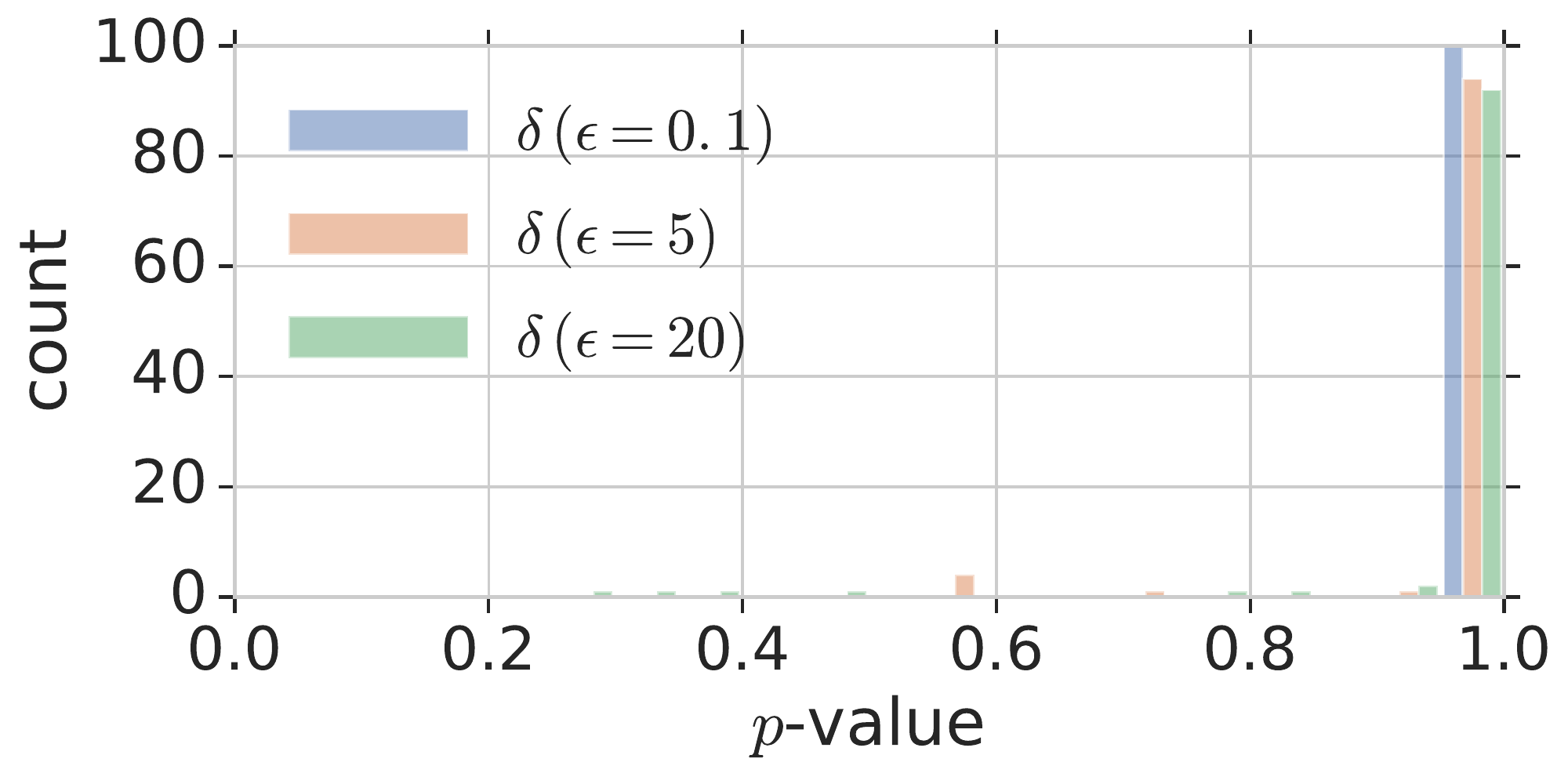} & \mbox{} &
  \includegraphics[width=0.42\textwidth]{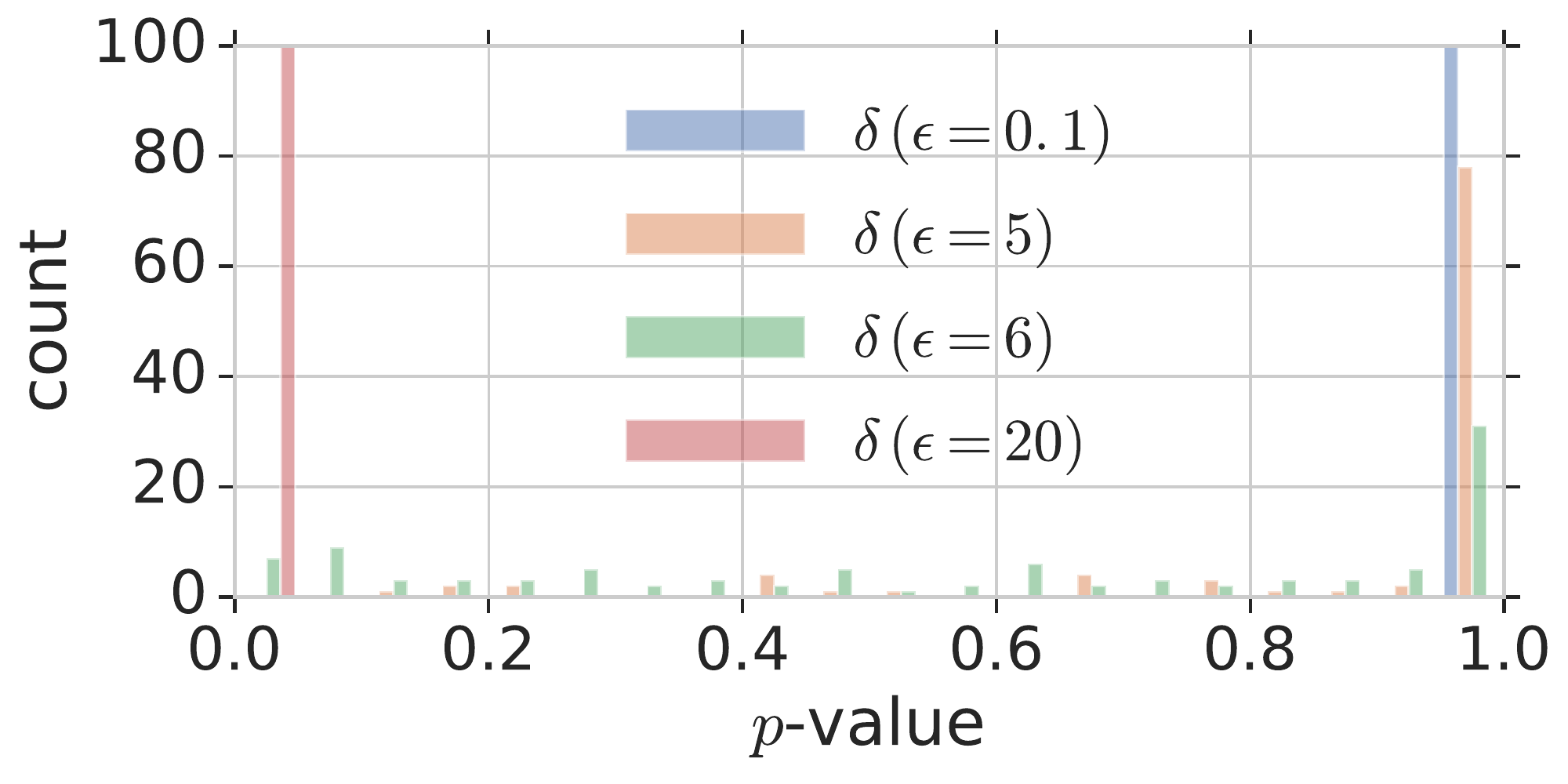} \\
  \end{tabular}
  \captionsetup{font=small}
  \caption{Risk and overfitting metrics for a synthetic problem with linear classifiers as a function of the perturbation strengths $\epsilon$ (log scale). Left: unbiased model tested on a large, independent test set (in this case $\hR_S(f) \approx \hR_g(f) \approx R(f)$); right: trained model overfitted to the test set ($\hR_S(f) \le \hR_g(f)$ while both are smaller than $R(f)$). 
    \emph{First row}: Average $p$-value $\delta$ for the pairwise independence test with over $100$ runs ($N=1$) or the $N$-model independence test ($N > 1$). The bounds plotted are either empirical $95\%$ two-sided confidence intervals ($N \le 2$) or ranges between minimum and maximum value ($N=10, 25$).
    \emph{Second row}: Empirical two-sided $97.5\%$  confidence intervals for the empirical test error rate $\hR_S(f)$ and the adversarial risk estimate $\hR_g(f)$. On the left, $R(f) \approx \hR_S(f)$, while $R(f)$ is shown separately on the right.
    \emph{Third row}: Average densities (Radon-Nikodym derivatives) for originally misclassified points and for the new data points obtained by successful adversarial transformations (with empirical 97.5\% two-sided confidence intervals).
    \emph{Fourth row}:   The empirical test error rate $\hR_S(f)$ and the adversarial risk estimate $\hR_g(f)$ for a single realization with $97.5\%$ two-sided confidence intervals computed from Bernstein's inequality, the adversarial error rate $\hR_{S'}(f)$, and the expected error $R(f)$ (on the right, on the left $R(f) \approx \hR_S(f)$).
    \emph{Fifth row}: Histograms of $p$-values for selected $\epsilon$ values over $100$ runs.
 \label{fig:linear-model}}
\end{figure}

The results of the two experiments are shown in \Cref{fig:linear-model}, plotted against different perturbation strengths: the left column corresponds to the first experiment while the right column to the second.
The first row presents the $p$-values for rejecting the independence hypothesis, calculated by repeating the experiment (sampling data and training the classifier) $100$ times and applying the single-model (\Cref{sec:detection}, labelled as $N=1$ in the plots) and $N$-model (\Cref{sec:detection-randomised}, labelled as $N=2, 10, 25, 100$ in the plots) independence test, and taking the average over models (or model sets of size $N$) for each $\epsilon$. We also plot empirical 95\% two-sided confidence intervals ($N \le 2$) or, due to limited number of $p$-values available after dividing 100 runs into disjoint bins of size $N \ge 10$, ranges between minimum and maximum value ($N=10, 25$). For all methods of detecting dependence, it can be seen that for the independent case the test is correctly not able to reject the independence hypothesis (the average $p$-value is very close to $1$, although in some runs it can drop to as low as $0.5$).
On the other hand, for $10 \le \epsilon \le 50$, the non-independent model failed the independence test at confidence level $1 - \delta \approx 100\%$, hence, in this range of $\epsilon$ our independence test reliably detects overfitting.

In fact, it is easy to argue that our test should only work for a limited range of $\epsilon$, that is, it should not reject independence for too small or too large values of $\epsilon$. 
First we consider the case of small $\epsilon$ values. Notice that except for points $g(x)$ $\epsilon$-close (in $L_2$-norm) to the true decision boundary or the decision boundary of $f$, $g(x)$ is invertible: if $g(x)$ is correctly classified and is $\epsilon$-away from the true decision boundary, there is exactly one point, $x$, which is translated to $g(x)$, while if $g(x)$ is incorrectly classified and $\epsilon$-away from the decision boundary of $f$, no translation leads to $g(x)$ and $x=g(x)$; any other points are $\epsilon$-close to the decision boundary of either $f$ or $f^*$. Thus, since $\rho$ is bounded, $g(x)$ is invertible on a set of at least $1-O(\epsilon)$ probability (according to $\rho$). When $\epsilon \to 0$, $g(x) \to x$, and so $\rho(g(x)) \to \rho(x)$ for all points $x$ with $|x_1| \neq 0.025$ (since $\rho$ is continuous in all such $x$), implying $h_g(g(x)) \approx 1$ on these points. It also follows that $L(f,x) \neq L(f,g(x))$ can only happen to a set of points with an $O(\epsilon)$ $\rho$-probability.
This means that $L(f,g(x))h_g(g(x)) \approx L(f,x)$ on a set of $1-O(\epsilon)$ $\rho$-probability, and for these points $T_g(x)=L(f,g(x)) h_g(g(x)) - L(f,x) \approx 0$.
Thus, $T_g(X) \approx 0$ with $\rho$-probability $1-O(\epsilon)$. Unless the test set $S$ is concentrated in large part on the set of remaining points with $O(\epsilon)$ $\rho$-probability, the test statistic $|T_{S,g}(f)|= O(\epsilon)$ with high probability and our method will not reject the independence hypothesis for $\epsilon \to 0$.

When $\epsilon$ is large ($\epsilon \to \infty$), notice that for any point $x$ with non-vanishing probability (i.e., with $\rho(x)>c$ for some $c>0$), if $g(x) \neq x$ than $\rho(g(x)) \approx 0$. Therefore, for such an $x$, if $L(f,x)=0$ and $L(f,g(x))=1$, $h_g(g(x))=\rho(g(x))/(\rho(x)+\rho(g(x))) \approx 0$, and so $T_g(x) \approx 0$ (if $L(f,g(x))=0$, we trivially have $T_g(x)=0$). If $L(f,x)=1$, we have $g(x)=x$. If $g$ is invertible at $x$ then $h_g(x)=1$ and  $T_g(x)=0$. If $g$ is not invertible, then there is another $x'$ such that $g(x')=x$; however, if $\rho(x)>c$ then $\rho(x') \approx 0$ (since $\epsilon$ is large), and so $h_g(g(x)) = \rho(x)/(\rho(x)+\rho(x')) \approx 1$, giving $T_g(x) \approx 0$.
Therefore, for large $\epsilon$, $T_g(X) \approx 0$ with high probability (i.e., for points with $\rho(x)>c$), so the independence hypothesis will not be rejected with high probability.

To better understand the behavior of the test, the second row of \Cref{fig:linear-model} shows the empirical test error rate $\hR_S(f)$, the (unadjusted) adversarial error rate $\hR_{S'}(f)$, and the adversarial risk estimate $\hR_g(f)$, together with their confidence intervals. For the non-independent model, we also show the expected error $R(f)$ (estimated over a large independent test set), while it is omitted for the independent model where it approximately coincides with both $\hR_S(f)$ and $\hR_g(f)$.
While the reweighted adversarial error estimate $\hR_g(f)$ remains the same for all perturbations in case of an independent test set (left column), the adversarial error rate $\hR_{S'}(f)$ varies a lot for both the dependent and independent test sets. For example, in the case when the test samples and the model $f$ are not independent, it undershoots the true error for $\epsilon < 10$ and overshoots it for larger perturbations. For very large perturbations ($\epsilon$ close to $100$), the behavior of $\hR_{S'}(f)$  depends on the model $f$: in the independent case $\hR_{S'}(f)$ decreases back to $\hR_S(f)$ because such large perturbations increasingly often change the true label of the original example, so less and less adversarial points are generated.  In the case when the data and the model are not independent (right column), the adversarial perturbations are almost always successful (i.e., lead to a valid adversarial example for most originally correctly classified points), yielding an adversarial error rate close to one for large enough perturbations. This is because the decision boundary of $f$ is almost orthogonal to the true decision boundary, and so the adversarial perturbations are parallel with the true boundary, almost never changing the true label of a point. 

The plots of the densities (Radon-Nikodym derivatives), given in the third row of \Cref{fig:linear-model}, show how the change in their values compensate the increase of the adversarial error rate $\hR_{S'}(f)$: in the independent case, the effect is completely eliminated yielding an unbiased adversarial error estimate $\hR_g(f)$, which is essentially constant over the whole range of $\epsilon$ (as shown in the first row), while in the non-independent case the similar densities do not bring back the adversarial error rate $\hR_{S'}(f)$ to the test error rate $\hR_S(f)$, allowing the test to detect overfitting. Note that the densities exhibit similar trends (and values) in both cases, driven by the dependence of typical values of the $\rho(x) / \rho(g(x))$ ratio on the perturbation strength $\epsilon$ for originally misclassifed points ($L(f,x)=1$) and for successful adversarial examples (i.e., $L(f,x)=0$ and $L(f,g(x))=1$). 

To compare the behavior of our improved, pairwise test and the basic version, the fourth row of \Cref{fig:linear-model} depicts a single realization of the experiments where the $97.5\%$ confidence intervals (as computed from Bernstein's inequality) are shown for the estimates. For the independent case, the confidence intervals of $\hR_S(f)$ and $\hR_g(f)$ overlap for all $\epsilon$, and thus the basic test is not able to detect overfitting. In the non-independent case, the confidence intervals overlap for $\epsilon=10$ and $\epsilon=75$, thus the basic test is not able to detect overfitting with at a $95\%$ confidence level, while the improved test (second row) is able to reject the independence hypothesis for these $\epsilon$ values at the same confidence level.

Finally, in the fifth row of \Cref{fig:linear-model} we plotted the histograms of the empirical distribution of $p$-values for both models, over 100 independent runs (between the runs, all the data was regenerated and the models were retrained).
For $\epsilon=0.1, 5, 20$, they concentrate heavily on either $\delta=0$ or $\delta=1$, and have very thin tails extending far towards the opposite end of the $[0,1]$ interval. This explains the surprisingly wide $95\%$ confidence intervals for $p$-values plotted in the first row. In particular, the fact that some $p$-values for the independent model are as low as $0.5$ does not mean the independence test is not reliable, because almost all calculated $\delta$ values are close or equal to $1$, and the few outliers are a combined consequence of the finite sample size and the effectiveness of the AEG. 
The additional $\epsilon=6$ histogram for the non-independent model illustrates a regime which is in between the single-model pairwise test (\Cref{sec:detection}) completely failing to reject the independence hypothesis and clearly rejecting it.

\begin{figure}[tbp]
  \centering\begin{tabular}{ccc}
  \includegraphics[width=0.42\textwidth]{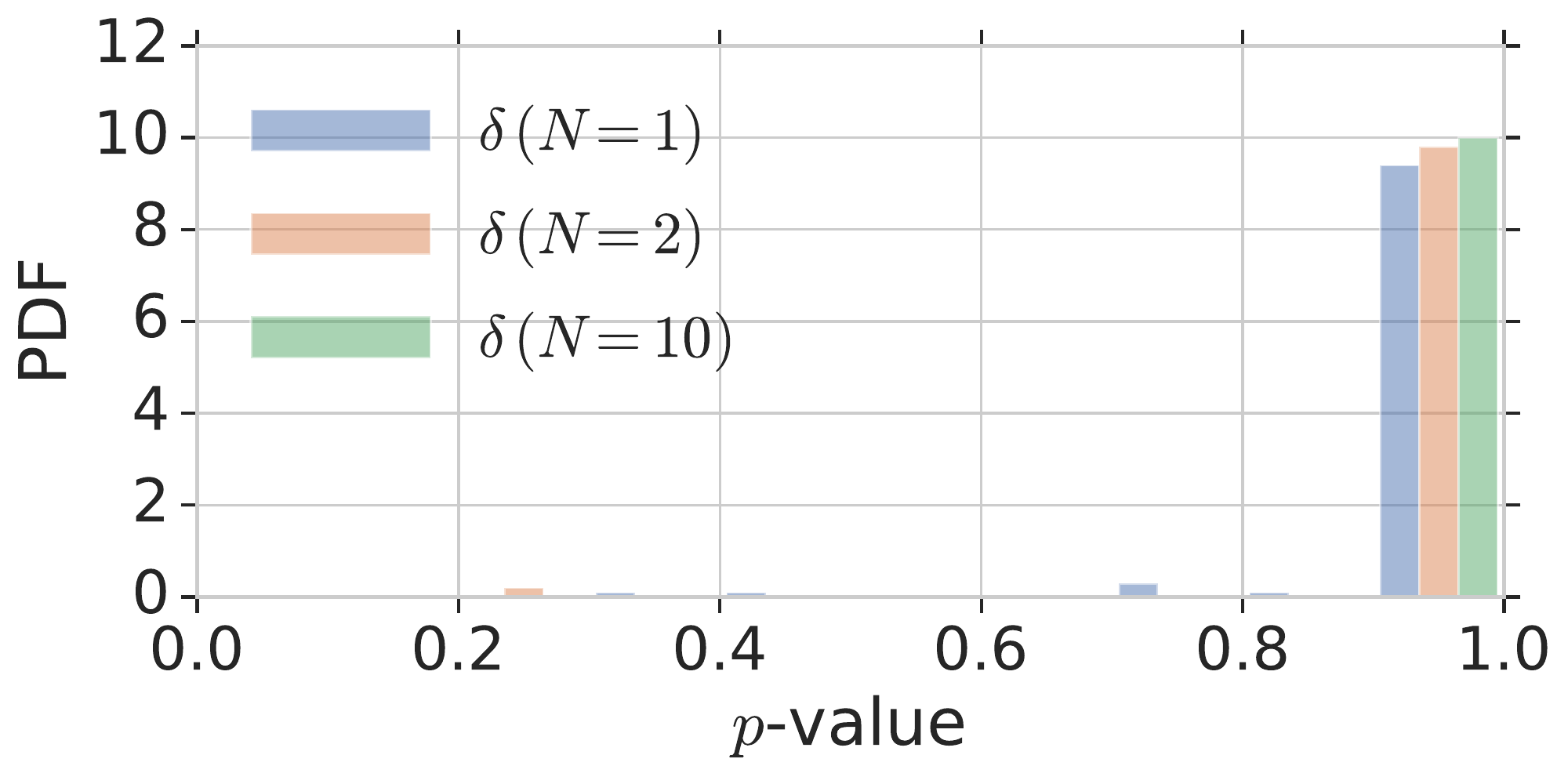}  & \mbox{\hspace{0.8cm}} &
  \includegraphics[width=0.42\textwidth]{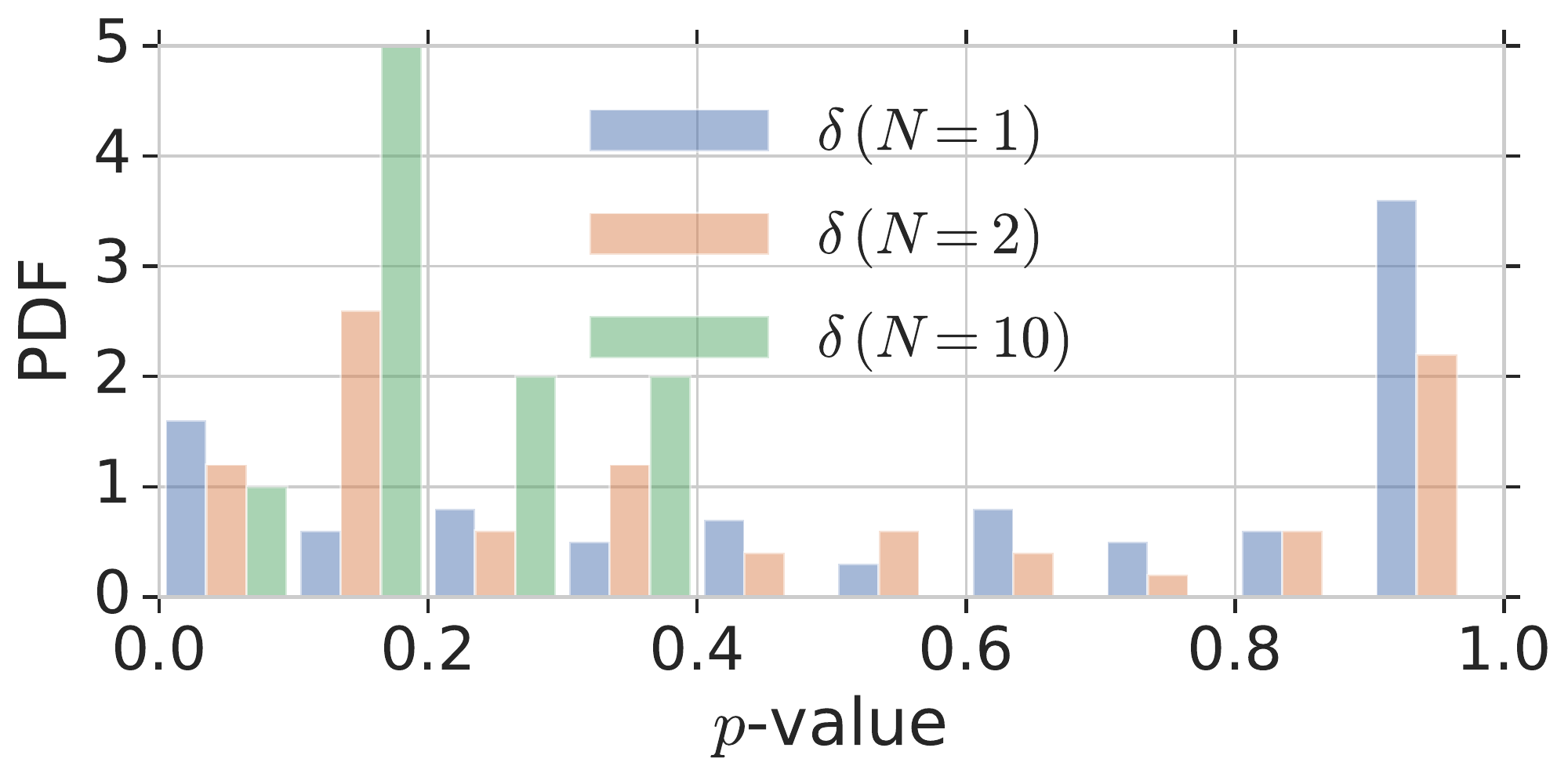} \\
  independent model, $\epsilon=10$ & \mbox{} & non-independent model, $\epsilon=6$
  \end{tabular}
  \captionsetup{font=small}
  \caption{Histograms of $p$-values from $N$-model ($N=1,2,10$) independence tests for both synthetic models and selected $\epsilon$ values, over $100$ runs.
 \label{fig:linear-model-multi}}
\end{figure}
To verify experimentally whether the $N$-model independence test can be a more powerful detector of overfitting than the single-model version, in \Cref{fig:linear-model-multi} (right panel) we plotted $p$-value histograms for $N=1,2,10$ for the intermediate AEG strength $\epsilon=6$ applied to the non-independent model over 100 training runs. Indeed, as $N$ increases, the concentration of $p$-values around in the low ($\delta \le 0.2$) range increases. For $N > 10$ we did not have enough values to plot a histogram: for $N = 25$ we obtained $\delta = 0.1851, 0.1599, 0.0661$ and $0.1941$, while for $N=100$ the $p$-value is $0.1153$. The increase of the test power becomes apparent when we compare the last value with the mean of $p$-values obtained by testing every training run separately, equal $0.5984$, and the median $0.6385$.

For comparison, we also plotted in \Cref{fig:linear-model-multi} (left panel) the corresponding histograms for the independent model and a slightly higher attack strength, $\epsilon=10$, at which the independence tests fails for the overfitted model even without averaging (see \Cref{fig:linear-model}, first row, right panel). The histograms are all clustered in the $\delta$ region close to 1, indicating that the $N$-model test is not overly pessimistic.

\section{Translational AEGs for image classification models}
\label{app:aeg-image-transl}

For image classification we consider two translation variants that are used in constructing a translational AEG.
For every correctly classified image $x$, we consider translations from $\V_\e$ (for some $\e$), choosing $g(x)$ from the set $G(x)=\{ \tau_v(x) : v \in \V_\e \} \cup \{x\}$. If all translations result in correctly classified examples, we set $g(x) = x$. Otherwise, we use one of two possible ways to select $g(x)$ (and we call the resulting points successful adversarial examples):
\begin{itemize}
\item \emph{Strongest perturbation:} Assuming the number of classes is $K$, let $l(f,x) \in \mathbb{R}^K$ denote the vector of the $K$ class logits calculated by the model $f$ for image $x$, and let
$l_\text{exc}(f,x) = \max_{0 \le i < K} l_i(f,x) - l_y(f, x)$.
We define $$g_\text{strongest}(x) = \argmax_{x' \in G(x)} l_\text{exc}(f,x'),$$ with ties broken deterministically by choosing the first translation from the candidate set, going top to bottom and left to right in row-major order. Thus, here we seek a non-identical ``neighbor'' that causes the classifier to err the most, reachable from $x$ by translations within a maximum range $\e$.
\item \emph{Nearest misclassified neighbor:} Here we aim to find the nearest image in $G(x)$ that is misclassified. That is, letting 
$d(x,x') = \|v\|_2$ if $x'=\tau_v(x)$ and $\infty$ otherwise, we define
\[
g_\text{nearest}(x) := \argmin_{x' \in G(x), L(f,x') = 1} d(x,x')
\]
with ties broken deterministically as above.
\end{itemize}
The two perturbation variants are successful on exactly the same set of images, hence they lead to the same adversarial error rates $\hR_{S'}(f)$. However, they are characterized by different values of the density $h_g$ and, consequently, yield different adversarial risk estimates $\hR_g(f)$ and associated $p$-values for the independence test. The main difference between them is that the ``strongest'' version is more likely to map multiple images to the same adversarial example, thus decreasing the densities for successful adversarial examples and, counterintuitively, increasing them for originally misclassified points (as their neighbors are less likely to be mapped to these points).

To better see the effect of adversarial perturbations, we also consider two random baselines that do not take into account the success of a translation in generating misclassified points: 
$g_\text{random}(x)$ is chosen uniformly at random from $G(x) \setminus \{x\}$, and $g_\text{random2}(x)$ is chosen uniformly at random from $G(x)$.

\subsection{Maximum translations}
\label{app:maxtranslation}

\begin{figure}[h]
  \begin{center}
    \includegraphics[width=0.4\columnwidth]{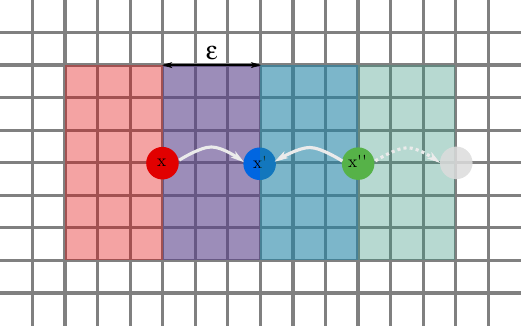}
  \end{center}
  \captionsetup{font=footnotesize}
  \caption{Image translations which need to be considered for a translational AEG with $\epsilon=3$. The red, blue and green balls represent the center of the original image $x$, adversarial example $x' = g(x)$ and another image $x''$ contributing to $\rho_g(x')$, respectively, while the semi-translucent squares of corresponding colors represent the possible translations which need to be considered for each of $x$, $x'$ and $x''$. Solid light grey arrows represent the relationships $x'=g(x)$ and $x'=g(x'')$. Finally, the dashed arrow and the semi-translucent grey ball represent an alternative mapping, which has to be ruled out while calculating the value of $g(x'')$ and, consequently, of $h_g(x')$. It is easy to see that the colored squares (which contain the translations needing to be evaluated) extend as far as $3\epsilon$ from the original image $x$.}
  \label{fig:grid}
\end{figure}

In practice, translating an image is not always simple, as the new image has to be padded with new pixels. When (central) crops of a larger image are used (as is typical for ImageNet classifiers), translations can easily be implemented as long as the resulting new cropping window stays within the original image boundaries.
Even if an image can be translated by a vector $v$, this limits our ability to compute $h_g(x')$ for the adversarial image $x'$ by \eqref{eq:hg} or \eqref{eq:hgnd} for $g_\text{strongest}$ or $g_\text{nearest}$. Indeed,  if an image $x$ is shifted by $v \in \V_\e$ to generate adversarial example $x'$, we need to examine translations of $x'$ with vectors in $\V_\e$ to find the neighbors $x''$ of $x'$ potentially contributing to $n(x')$ when computing $h_g(x')$. Finally we need to consider translations of $x''$ with vectors in $\V_\e$ to determine the exact value they contribute, that is, to compute the exact probabilities in \eqref{eq:hgnd} (see \Cref{fig:grid} for an illustration).
Thus, to be able to compute the density $h_g$ for the adversarial points obtained by translations from $\V_\e$, we might need to be able to perform translations within $\V_{3\e}$.

\end{document}